\documentclass[preprint, prX]{revtex4}

\usepackage{amsmath}    
\usepackage{graphicx}   
\usepackage{verbatim}   
\usepackage{color}      
\usepackage{subfigure}  
\usepackage{hyperref}   
\usepackage{amssymb}    
\usepackage{multirow}
\usepackage{epsfig}
\usepackage{graphics,graphicx}
\usepackage{setspace}
\usepackage{url}
\usepackage{algorithm,algorithmic}

\begin{document}

\begin{abstract}
Evolutionary algorithms are good general problem solver but suffer from a lack of domain specific knowledge.
However, the problem specific knowledge can be added to evolutionary algorithms by hybridizing. Interestingly,
all the elements of the evolutionary algorithms can be hybridized. In this chapter, the hybridization of the three
elements of the evolutionary algorithms is discussed: the objective function, the survivor selection operator and
the parameter settings. As an objective function, the existing heuristic function that construct the solution of the
problem in traditional way is used. However, this function is embedded into the evolutionary algorithm that serves
as a generator of new solutions. In addition, the objective function is improved by local search heuristics. The
new neutral selection operator has been developed that is capable to deal with neutral solutions, i.e. solutions that
have the different representation but expose the equal values of objective function. The aim of this operator is to
directs the evolutionary search into a new undiscovered regions of the search space. To avoid of wrong setting of
parameters that control the behavior of the evolutionary algorithm, the self-adaptation is used. Finally, such hybrid
self-adaptive evolutionary algorithm is applied to the two real-world NP-hard problems: the graph 3-coloring and
the optimization of markers in the clothing industry. Extensive experiments shown that these hybridization improves
the results of the evolutionary algorithms a lot. Furthermore, the impact of the particular hybridizations is analyzed
in details as well.

\textit{To cite paper as follows: Iztok Fister, Marjan Mernik and Janez Brest (2011). Hybridization of Evolutionary Algorithms, Evolutionary Algorithms, Eisuke Kita (Ed.), ISBN: 978-953-307-171-8, InTech, Available from: http://www.intechopen.com/books/evolutionary-algorithms/hybridization-of-evolutionary-algorithms
}

\end{abstract}

\title{Hybridization of Evolutionary Algorithms}

\author{Iztok Fister}
\altaffiliation{University of Maribor, Faculty of electrical engineering and computer science
Smetanova 17, 2000 Maribor}
\email{iztok.fister@uni-mb.si}

\author{Marjan Mernik}
\altaffiliation{University of Maribor, Faculty of electrical engineering and computer science
Smetanova 17, 2000 Maribor}
\email{marjan.mernik@uni-mb.si}

\author{Janez Brest}
\altaffiliation{University of Maribor, Faculty of electrical engineering and computer science
Smetanova 17, 2000 Maribor}
\email{janez.brest@uni-mb.si}

\maketitle

\section{Introduction}
Evolutionary algorithms are a type of general problem solvers that can be applied to many difficult optimization problems. Because of their generality, these algorithms act similarly like Swiss Army knife \citep{book:michalewicz2004} that is a handy set of tools that can be used to address a variety of tasks. In general, a definite task can be performed better with an associated special tool. However, in the absence of this tool, the Swiss Army knife may be more suitable as a substitute. For example, to cut a piece of bread the kitchen knife is more suitable, but when traveling the Swiss Army knife is fine.

Similarly, when a problem to be solved from a domain where the problem-specific knowledge is absent evolutionary algorithms can be successfully applied. Evolutionary algorithms are easy to implement and often provide adequate solutions. An origin of these algorithms is found in the Darwian principles of natural selection \citep{book:darwin1859}. In accordance with these principles, only the fittest individuals can survive in the struggle for existence and reproduce their good characteristics into next generation.

\begin{figure}[htb]	
\centering
\includegraphics[width=8cm]{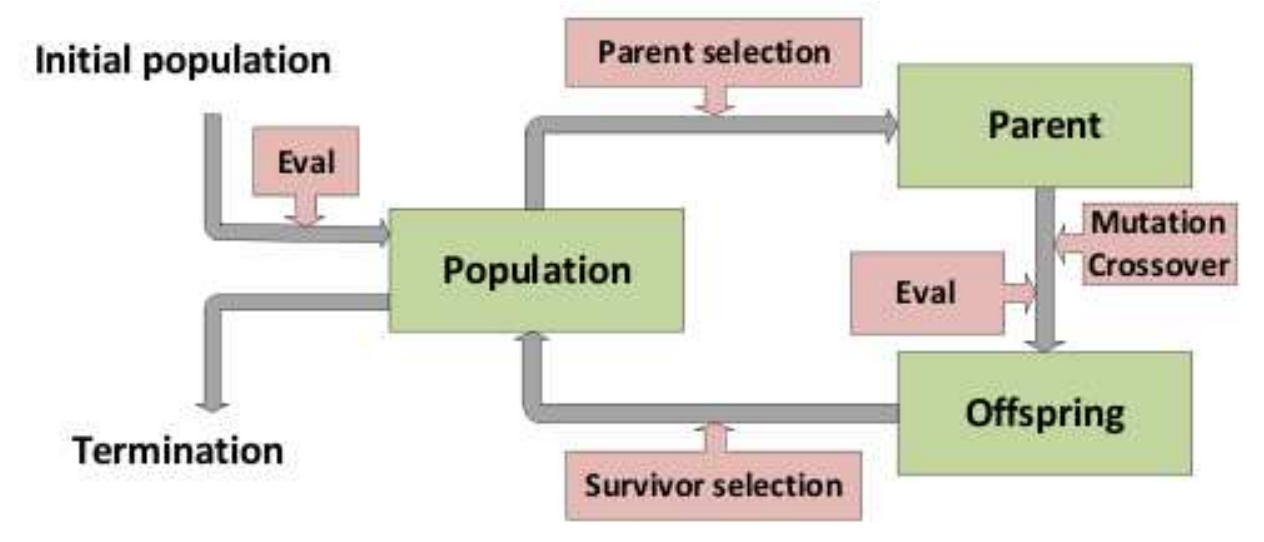} 
\caption{Scheme of Evolutionary Algorithms}
\label{pic:1}
\end{figure}

As illustrated in Fig.~\ref{pic:1}, evolutionary algorithms operate with the population of solutions. At first, the solution needs to be defined within an evolutionary algorithm. Usually, this definition cannot be described in the original problem context directly. In contrast, the solution is defined by data structures that describe the original problem context indirectly and thus, determine the search space within an evolutionary search (optimization process). There exists the analogy in the nature, where the genotype encodes the phenotype, as well. Consequently, a genotype-phenotype mapping determines how the genotypic representation is mapped to the phenotypic property. In other words, the phenotypic property determines the solution in original problem context. Before an evolutionary process actually starts, the initial population needs to be generated. The initial population is generated most often randomly. A basis of an evolutionary algorithm represents an evolutionary search in which the selected solutions undergo an operation of reproduction, i.e., a crossover and a mutation. As a result, new candidate solutions (offsprings) are produced that compete, according to their fitness, with old ones for a place in the next generation. The fitness is evaluated by an evaluation function (also called fitness function) that defines requirements of the optimization (minimization or maximization of the fitness function). In this study, the minimization of the fitness function is considered. As the population evolves solutions becomes fitter and fitter. Finally, the evolutionary search can be iterated until a solution with sufficient quality (fitness) is found or the predefined number of generations is reached \citep{book:eiben2003}. Note that some steps in Fig.~\ref{pic:1} can be omitted (e.g., mutation, survivor selection).

An evolutionary search is categorized by two terms: exploration and exploitation. The former term is connected with a discovering of the new solutions, while the later with a search in the vicinity of knowing good solutions \citep{book:eiben2003,art:mernik2009}. Both terms, however, interweave each other in the evolutionary search. The evolutionary search acts correctly when a sufficient diversity of population is present. The population diversity can be measured differently: the number of different fitness values, the number of different genotypes, the number of different phenotypes, entropy, etc. The higher the population diversity, the better exploration can be expected. Losing of population diversity can lead to the premature convergence.

Exploration and exploitation of evolutionary algorithms are controlled by the control parameters, for instance the population size, the probability of mutation $p_{m}$, the probability of crossover $p_{c}$, and the tournament size. To avoid a wrong setting of these, the control parameters can be embedded into the genotype of individuals together with problem variables and undergo through evolutionary operations. This idea is exploited by a self-adaptation. The performance of a self-adaptive evolutionary algorithm depends on the characteristics of population distribution that directs the evolutionary search towards appropriate regions of the search space \citep{inbook:lobo2007}.
\cite{art:igel2003}, however, widened the notion of self-adaptation with a generalized concept of self-adaptation. This concept relies on the neutral theory of molecular evolution \citep{art:kimura1968}. Regarding this theory, the most mutations on molecular level are selection neutral and therefore, cannot have any impact on fitness of individual. Consequently, the major part of evolutionary changes are not result of natural selection but result of random genetic drift that acts on neutral allele. An neutral allele is one or more forms of a particular gene that has no impact on fitness of individual \citep{book:hamilton2009}. In contrast to natural selection, the random genetic drift is a whole stochastic process that is caused by sampling error and affects the frequency of mutated allele. On basis of this theory Igel and Toussaint ascertain that the neutral genotype-phenotype mapping is not injective. That is, more genotypes can be mapped into the same phenotype. By self-adaptation, a neutral part of genotype (problem variables) that determines the phenotype enables discovering the search space independent of the phenotypic variations. On the other hand, the rest part of genotype (control parameters) determines the strategy of discovering the search space and therefore, influences the exploration distribution.

Although evolutionary algorithms can be applied to many real-world optimization problems their performance is still subject of the No Free Lunch (NFL) theorem \citep{art:wolpert1997}. According to this theorem any two algorithms are equivalent, when their performance is compared across all possible problems. Fortunately, the NFL theorem can be circumvented for a given problem by a hybridization that incorporates the problem specific knowledge into evolutionary algorithms.

\begin{figure}[htb]	
\centering
\includegraphics[width=10cm]{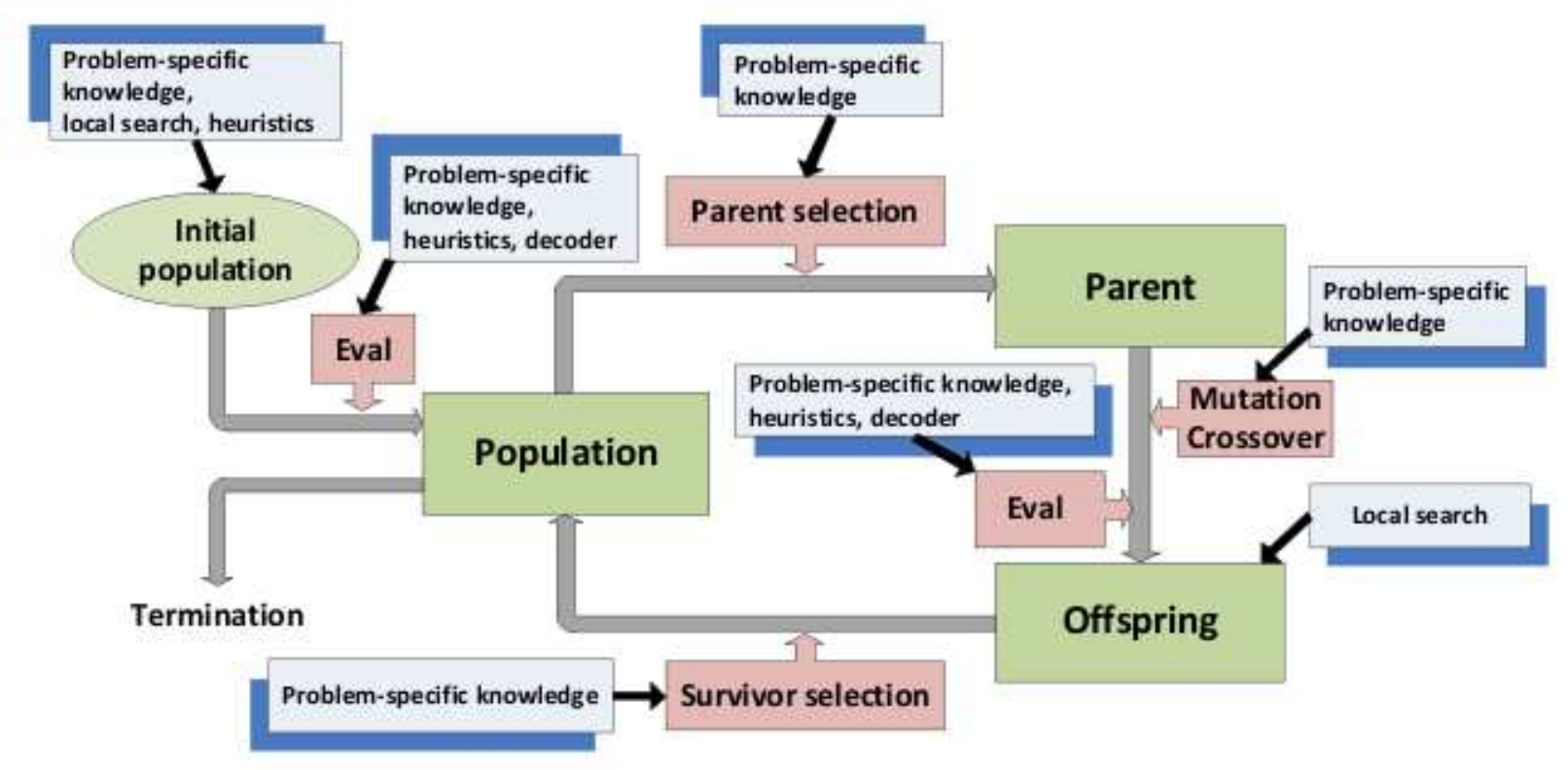} 
\caption{Hybridization of Evolutionary Algorithms}
\label{pic:2}
\end{figure}

In Fig. 2 some possibilities to hybridize evolutionary algorithms are illustrated. At first, the initial population can be generated by incorporating solutions of existing algorithms or by using heuristics, local search, etc. In addition, the local search can be applied to the population of offsprings. Actually, the evolutionary algorithm hybridized with local search is called a memetic algorithm as well \citep{inbook:moscato1999,art:wilfried2010}. Evolutionary operators (mutation, crossover, parent and survivor selection) can incorporate problem-specific knowledge or apply the operators from other algorithms. Finally, a fitness function offers the most possibilities for a hybridization because it can be used as decoder that decodes the indirect represented genotype into feasible solution. By this mapping, however, the problem specific knowledge or known heuristics can be incorporated to the problem solver.

In this chapter the hybrid self-adaptive evolutionary algorithm (HSA-EA) is presented that is hybridized with:
\begin{itemize}
  \item construction heuristic,
  \item local search,
  \item neutral survivor selection, and
  \item heuristic initialization procedure.
\end{itemize}
This algorithm acts as meta-heuristic, where the down-level evolutionary algorithm is used as generator of new solutions, while for the upper-level construction of the solutions a traditional heuristic is applied. This construction heuristic represents the hybridization of evaluation function. Each generated solution is improved by the local search heuristics. This evolutionary algorithm supports an existence of neutral solutions, i.e., solutions with equal values of a fitness function but different genotype representation. Such solutions can be arisen often in matured generations of evolutionary process and are subject of neutral survivor selection. This selection operator models oneself upon a neutral theory of molecular evolution \citep{art:kimura1968} and tries to direct the evolutionary search to new, undiscovered regions of search space. In fact, the neutral survivor selection represents hybridization of evolutionary operators, in this case, the survivor selection operator. The hybrid self-adaptive evolutionary algorithm can be used especially for solving of the hardest combinatorial optimization problems \citep{art:fister2010}.

The chapter is further organized as follows. In the Sect. 2 the self-adaptation in evolutionary algorithms is discussed. There, the connection between neutrality and self-adaptation is explained. Sect. 3 describes hybridization elements of the self-adaptive evolutionary algorithm. Sect. 4 introduces the implementations of hybrid self-adaptive evolutionary algorithm for graph 3-coloring in details. Performances of this algorithm are substantiated with extensive collection of results. The chapter is concluded with summarization of the performed work and announcement of the possibilities for the further work.


\section{The Self-adaptive Evolutionary Algorithms}

Optimization is a dynamical process, therefore, the values of parameters that are set at initialization become worse through the run. The necessity to adapt control parameters during the runs of evolutionary algorithms born an idea of self-adaptation \citep{book:holland1992}, where some control parameters are embedded into genotype. This genotype undergoes effects of variation operators. Mostly, with the notion of self-adaptation Evolutionary Strategies \citep{book:beyer1998,book:rechenberg1973,inbook:schwefel1977} are connected that are used for solving continuous optimization problems. Typically, the problem variables in Evolutionary Strategies are represented as real-coded vector $y=(y_{1},\ldots,y_{n})$ that are embedded into genotype together with control parameters (mostly mutation parameters). These parameters determine mutation strengths $\sigma$ that must be greater than zero. Usually, the mutation strengths are assigned to each problem variable. In that case, the uncorrelated mutation with $n$ step sizes is obtained \citep{book:eiben2003}. Here, the candidate solution is represented as $(y_{1},\ldots,y_{n},\sigma_{1},\ldots,\sigma_{n})$. The mutation is now specified as follows:
\begin{equation}
\label{eq:es-1}
 \sigma_{i}^{'}=\sigma_{i}\cdot \textnormal{exp}(\tau^{'}\cdot \textnormal{N}(0,1)+\tau\cdot \textnormal{N}_{i}(0,1)),
\end{equation}
\begin{equation}
\label{eq:es-2}
 y_{i}^{'}=y_{i}+\sigma _{i}^{'}\cdot \textnormal{N}_{i}(0,1),
\end{equation}
where $\tau ^{'}\propto 1/\sqrt{2\cdot n}$ and $\tau \propto 1/\sqrt{2\cdot \sqrt{n}}$ denote the learning rates. To keep the mutation strengths $\sigma_{i}$ greater than zero, the following rule is used
\begin{equation}
\label{eq:eps}
 \sigma _{i}<\varepsilon _{0}\Rightarrow \sigma _{i}=\varepsilon _{0}.
\end{equation}
Frequently, a crossover operator is used in the self-adaptive Evolutionary Strategies. This operator from two parents forms one offsprings. Typically, a discrete and arithmetic crossover is used. The former, from among the values of two parents $x_{i}$ and $y_{i}$ that are located on $i$-th position, selects the value of offspring $z_{i}$ randomly. The later calculates the value of offspring $z_{i}$ from the values of two parents $x_{i}$ and $y_{i}$ that are located on $i$-th position according to the following equation:
\begin{equation}
\label{eq:xover}
 z_{i}=\alpha \cdot x_{i}+(1-\alpha)\cdot y_{i},
\end{equation}
where parameter $\alpha$ captures the values from interval $\alpha \in \left [ 0 \ldots 1 \right ]$. In the case of $\alpha =1/2$, the uniform arithmetic crossover is obtained.

The potential benefits of neutrality was subject of many researches in the biological science \citep{art:conrad1990,art:hynen1996,art:kimura1968}. At the same time, the growing interest for the usage of this knowledge in evolutionary computation was raised \citep{inbook:barnett1998,inproc:ebner2001}. \cite{inproc:toussaint2002} dealt with the non-injectivity of genotype-phenotype mapping that is the main characteristic of this mapping. That is, more genotypes can be mapped to the same phenotype. \cite{art:igel2003} pointed out that in the absence of an external control and with a constant genotype-phenotype mapping only neutral genetic variations can allow an adaptation of exploration distribution without changing the phenotypes in the population. However, the neutral genetic variations act on the genotype of parent but does not influence on the phenotype of offspring.

As a result, control parameters in evolutionary strategies represent a search strategy. The change of this strategy enables a discovery of new regions of the search space. The genotype, therefore, does not include only the information addressing its phenotype but the information about further discovering of the search space as well. In summary, the neutrality is not necessary redundant but it is prerequisite for self-adaptation. This concept is called the general concept of self-adaptation as well \citep{inbook:lobo2007}.

\section{How to hybridize the Self-adaptive Evolutionary Algorithms}

Evolutionary algorithms are a generic tool that can be used for solving many hard optimization problems. However, the solving of that problems showed that evolutionary algorithms are too problem-independent. Therefore, there are hybridized with several techniques and heuristics that are capable to incorporate problem-specific knowledge. \cite{inbook:grosan2007} identified mostly used hybrid architectures today as follows:
\begin{itemize}
  \item hybridization between two evolutionary algorithms \citep{art:grefenstette1986},
  \item neural network assisted evolutionary algorithm \citep{art:wang2005},
  \item fuzzy logic assisted evolutionary algorithm \citep{inbook:herrera1996,inproc:lee1993},
  \item particle swarm optimization assisted evolutionary algorithm \citep{inproc:eberhart1995,inproc:kennedy1995},
  \item ant colony optimization assisted evolutionary algorithm \citep{inbook:fleurent1994,art:tseng2005},
  \item bacterial foraging optimization assisted evolutionary algorithm \citep{art:kim2005,art:neppalli1996},
  \item hybridization between an evolutionary algorithm and other heuristics, like local search \citep{inbook:moscato1999}, tabu search \citep{art:galinier1999}, simulated annealing \citep{art:ganesh2004}, hill climbing \citep{book:koza2003}, dynamic programming \citep{inproc:doerr2009}, etc.
\end{itemize}

In general, successfully implementation of evolutionary algorithms for solving a given problem depends on incorporated problem-specific knowledge. As already mentioned before, all elements of evolutionary algorithms can be hybridized. Mostly, a hybridization addresses the following elements of evolutionary algorithms  \citep{book:michalewicz1992}:
\begin{itemize}
  \item initial population,
  \item genotype-phenotype mapping,
  \item evaluation function, and
  \item variation and selection operators.
\end{itemize}

First, problem-specific knowledge incorporated into heuristic procedures can be used for creating an initial population. Second, genotype-phenotype mapping is used by evolutionary algorithms, where the solutions are represented in an indirect way. In that cases, a constructing algorithm that maps the genotype representation into a corresponding phenotypic solution needs to be applied. This constructor can incorporate various heuristic or other problem-specific knowledge. Third, to improve the current solutions by an evaluation function, local search heuristics can be used. Finally, problem-specific knowledge can be exploited by heuristic variation and selection operators.

The mentioned hybridizations can be used to hybridize the self-adaptive evolutionary algorithms as well. In the rest of chapter, we propose three kinds of hybridizations that was employed to the proposed hybrid self-adaptive evolutionary algorithms:
\begin{itemize}
  \item the construction heuristics that can be used by the genotype-phenotype mapping,
  \item the local search heuristics that can be used by the evaluation function, and
  \item the neutral survivor selection that incorporates the problem-specific knowledge.
\end{itemize}
Because the initialization of initial population is problem dependent we omit it from our discussion.

\subsection{The Construction Heuristics}

Usually, evolutionary algorithms are used for problem solving, where a lot of experience and knowledge is accumulated in various heuristic algorithms. Typically, these algorithms work well on limited number of problems \citep{book:hoos2005}. On the other hand, evolutionary algorithms are a general method suitable to solve very different kinds of problems. In general, these algorithms are less efficient than heuristics specialized to solve the given problem. If we want to combine a benefit of both kind of algorithms then the evolutionary algorithm can be used for discovering new solutions that the heuristic exploits for building of new, probably better solutions. Construction heuristics build the solution of optimization problem incrementally, i.e., elements are added to a solution step by step (Algorithm~\ref{alg:const}).

\begin{algorithm}[t!]
\caption{The construction heuristic. $I$: task, $S$: solution.}
\label{alg:const}
\begin{algorithmic}[1]
\WHILE{\textbf{NOT} $final\_solution(y \in S)$}
    \STATE $add\_element\_to\_solution\_heuristicaly (y_{i} \in I,S)$;
\ENDWHILE
\end{algorithmic}
\end{algorithm}

\subsection{The Local Search}

A local search belongs to a class of improvement heuristics \citep{book:aarts1997}. In our case, main characteristic of these is that the current solution is taken and improved as long as improvements are perceived.

The local search is an iterative process of discovering points in the vicinity of current solution. If a better solution is found the current solution is replaced by it. A neighborhood of the current solution $y$ is defined as a set of solutions that can be reached using an unary operator $\mathcal{N}:S \rightarrow 2^{S}$ \citep{book:hoos2005}. In fact, each neighbor $y'$ in neighborhood $\mathcal{N}$ can be reached from current solution $y$ in $k$ strokes. Therefore, this neighborhood is called $k \textendash opt$ neighborhood of current solution $y$ as well. For example, let the binary represented solution $y$ and $1 \textendash opt$ operator on it are given. In that case, each of neighbors $\mathcal{N}(y)$ can be reached changing exactly one bit. The neighborhood of this operator is defined as
\begin{equation}
\label{eq:1-opt}
 \mathcal{N}_{1 \textendash opt}(y)=\{y' \in S|d_{H}(y,y')=1\},
\end{equation}
where $d_{H}$ denotes a Hamming distance of two binary vectors as follows
\begin{equation}
\label{eq:Hamming}
 d_{H}(y,y')=\sum _{i=1}^{n}(y_{i}\oplus y'_{i}),
\end{equation}
where operator $\oplus$ means $exclusive\ or$ operation. Essentially, the Hamming distance in Equation~\ref{eq:Hamming} is calculated by counting the number of different bits between vectors $y$ and $y'$. The $1 \textendash opt$ operator defines the set of feasible $1 \textendash opt$ strokes while the number of feasible $1 \textendash opt$ strokes determines the size of neighborhood.

\begin{algorithm}[t!]
\caption{The local search. $I$: task, $S$: solution.}
\label{alg:ls}
\begin{algorithmic}[1]
\STATE $generate\_initial\_solution(y \in S)$;
\REPEAT
    \STATE $find\_next\_neighbor (y' \in \mathcal{N}(y))$;
    \IF {$(f(y')<f(y))$}
        \STATE $y=y'$;
    \ENDIF
\UNTIL{$set\_of\_neighbor\_empty$;}
\end{algorithmic}
\end{algorithm}

As illustrated by Algorithm~\ref{alg:ls}, the local search can be described as follows \citep{book:michalewicz2004}:
\begin{itemize}
  \item The initial solution is generated that becomes the current solution (procedure $generate\_initial\_solution$).
  \item The current solution is transformed with $k \textendash opt$ strokes and the given solution $y'$ is evaluated (procedure $find\_next\_neighbor$).
  \item If the new solution $y'$ is better than the current $y$ the current solution is replaced. On the other hand, the current solution is kept.
  \item Lines 2 to 7 are repeated until the set of neighbors is not empty (procedure $set\_of\_neighbor\_empty$).
\end{itemize}

In summary, the $k \textendash opt$ operator represents a basic element of the local search from which depends how exhaustive the neighborhood will be discovered. Therefore, the problem-specific knowledge needs to be incorporated by building of the efficient operator.

\subsection{The Neutral Survivor Selection}

A genotype diversity is one of main prerequisites for the efficient self-adaptation. The smaller genotypic diversity causes that the population is crowded in the search space. As a result, the search space is exploited. On the other hand, the larger genotypic diversity causes that the population is more distributed within the search space and therefore, the search space is explored \citep{book:baeck1996}. Explicitly, the genotype diversity of population is maintained with a proposed neutral survivor selection that is inspired by the neutral theory of molecular evolution \citep{art:kimura1968}, where the neutral mutation determines to the individual three possible destinies, as follows:
\begin{itemize}
  \item the fittest individual can survive in the struggle for existence,
  \item the less fitter individual is eliminated by the natural selection,
  \item individual with the same fitness undergo an operation of genetic drift, where its survivor is dependent on a chance.
\end{itemize}

Each candidate solution represents a point in the search space. If the fitness value is assigned to each feasible solution then these form a fitness landscape that consists of peeks, valleys and plateaus \citep{inproc:wrights96}. In fact, the peaks in the fitness landscape represents points with higher fitness, the valleys points with the lower fitness while plateaus denotes regions, where the solutions are neutral \citep{inbook:stadler1995}. The concept of the fitness landscape plays an important role in evolutionary computation as well. Moreover, with its help behavior of evolutionary algorithms by solving the optimization problem can be understood. If on the search space we look from a standpoint of fitness landscape then the heuristical algorithm tries to navigate through this landscape with aim to discover the highest peeks in the landscape \citep{inbook:merz1999}.

However, to determine how distant one solution is from the other, some measure is needed. Which measure to use depends on a given problem. In the case of genetic algorithms, where we deal with the binary solutions, the Hamming distance (Equation~\ref{eq:Hamming}) can be used. When the solutions are represented as real-coded vectors an Euclidian distance is more appropriate. The Euclidian distance between two vectors $x$ and $y$ is expressed as follows:
\begin{equation}
\label{eq:Euclid}
 d_{E}(x,y)=\sqrt{\frac{1}{n}\cdot \sum_{i=1}^{n}(x_{i}-y_{i})^{2}},
\end{equation}
and measures the root of quadrat differences between elements of vectors $x$ and $y$. The main characteristics of fitness landscapes that have a great impact on the evolutionary search are the following \citep{inbook:merz1999}:
\begin{itemize}
  \item the fitness differences between neighboring points in the fitness landscape: to determine a ruggedness of the landscape, i.e., more rugged as the landscape, more difficultly the optimal solution can be found;
  \item the number of peaks (local optima) in the landscape: the higher the number of peaks, the more difficulty the evolutionary algorithms can direct the search to the optimal solution;
  \item how the local optima are distributed in the search space: to determine the distribution of the peeks in the fitness landscape;
  \item how the topology of the basins of attraction influences on the exit from the local optima: to determine how difficult the evolutionary search that gets stuck into local optima can find the exit from it and continue with the discovering of the search space;
  \item existence of the neutral networks: the solutions with the equal value of fitness represent a plateaus in the fitness landscape.
\end{itemize}

When the stochastic fitness function is used for evaluation of individuals the fitness landscape is changed over time. In this way, the dynamic landscape is obtained, where the concept of fitness landscape can be applied, first of all, to analyze the neutral networks that arise, typically, in the matured generations. To determine, how the solutions are dissipated over the search space some reference point is needed. For this reason, the current best solution $y^{*}$ in the population is used. This is added to the population of $\mu$ solutions.

An operation of the neutral survivor selection is divided into two phases. In the first phase, the evolutionary algorithm from the population of $\lambda$ offsprings finds a set of neutral solutions $N_{S}=\{y_{1}, \ldots ,y_{k}\}$ that represents the best solutions in the population of offsprings. If the neutral solutions are better than or equal to the reference, i.e. $f(y_{i}) \leq f(y^{*})$ for $i=1, \ldots ,k$, then reference solution $y^{*}$ is replaced with the neutral solution $y_{i} \in N_{S}$ that is the most faraway from reference solution according to the Equation 7. Thereby, it is expected that the evolutionary search is directed to the new, undiscovered region of the search space.
In the second phase, the updated reference solution $y^{*}$ is used to determine the next population of survivors. Therefore, all offsprings are ordered with regard to the ordering relation $\prec$ (read: is better than) as follows:
\begin{equation}
\label{eq:order}
 f(y_{1}) \prec \ldots \prec f(y_{i}) \prec f(y_{i+1}) \prec \ldots \prec f(y_{\lambda}),
\end{equation}

where the ordering relation $\prec$ is defined as

\begin{equation}
\label{eq:order2}
f(y_{i})\prec f(y_{i+1})\Rightarrow \left\{\begin{matrix}
f(y_{i})< f(y_{i+1}), \\
f(y_{i})= f(y_{i+1})\wedge (d(y_{i},y^{*})>d(y_{i+1},y^{*})).
\end{matrix}\right.
\end{equation}

Finally, for the next generation the evolutionary algorithm selects the best $\mu$ offsprings according to the Equation~\ref{eq:order}. These individuals capture the random positions in the next generation. Likewise the neutral theory of molecular evolution, the neutral survivor selection offers to the offsprings three possible outcomes, as follows. The best offsprings survive. Additionally, the offspring from the set of neutral solutions that is far away of reference solution can become the new reference solution. The less fitter offsprings are usually eliminated from the population. All other solutions, that can be neutral as well, can survive if they are ordered on the first $\mu$ positions regarding to Equation~\ref{eq:order}.









\section{The Hybrid Self-adaptive Evolutionary Algorithms in Practice}

In this section an implementation of the hybrid self-adaptive evolutionary algorithms (HSA-EA) for solving combinatorial optimization problems is represented. The implementation of this algorithm in practice consists of the following phases:
\begin{itemize}
  \item finding the best heuristic that solves the problem on a traditional way and adapting it to use by the self-adaptive evolutionary algorithm,
  \item defining the other elements of the self-adaptive evolutionary algorithm,
  \item defining the suitable local search heuristics, and
  \item including the neutral survivor selection.
\end{itemize}

\begin{figure}[htb]	
\centering
\includegraphics[width=8cm]{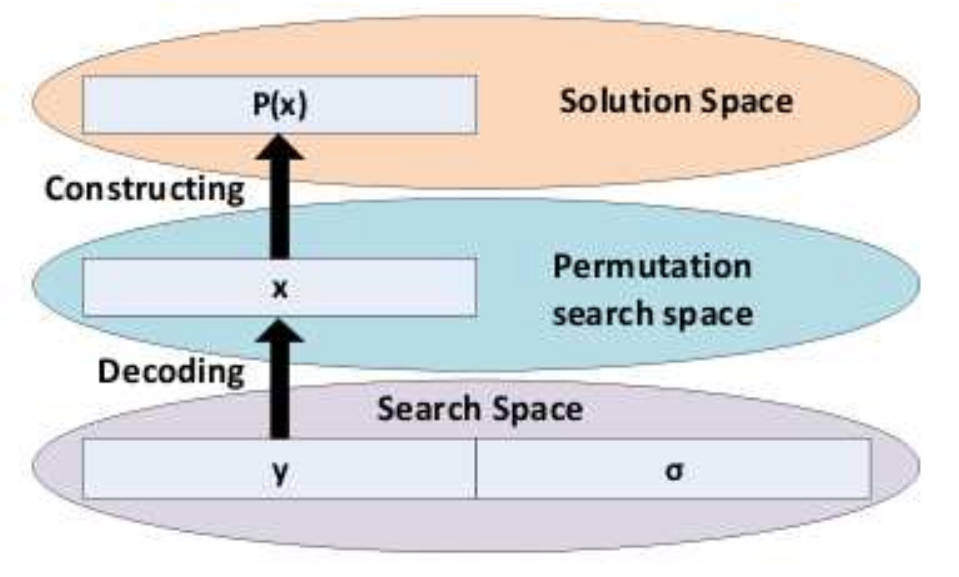} 
\caption{The genotype-phenotype mapping by hybrid self-adaptive evolutionary algorithm}
\label{pic:3}
\end{figure}

The main idea behind use of the construction heuristics in the HSA-EA is to exploit the knowledge accumulated in existing heuristics. Moreover, this knowledge is embedded into the evolutionary algorithm that is capable to discover the new solutions. To work simultaneously both algorithms need to operate with the same representation of solutions. If this is not a case a decoder can be used. The solutions are encoded by the evolutionary algorithm as the real-coded vectors and decoded before the construction of solutions. The whole task is performed in genotype-phenotype mapping that is illustrated in Fig.~\ref{pic:3}.

The genotype-phenotype mapping consists of two phases as follows:
\begin{itemize}
  \item decoding,
  \item constructing.
\end{itemize}
Evolutionary algorithms operate in genotypic search space, where each genotype consists of real-coded problem variables and control parameters. For encoded solution only the problem variables are taken. This solution is further decoded by decoder into a decoded solution that is appropriate for handling of a construction heuristic. Finally, the construction heuristic constructs the solution within the original problem context, i.e., problem solution space. This solution is evaluated by the suitable evaluation function.

The other elements of self-adaptive evolutionary algorithm consists of:
\begin{itemize}
  \item evaluation function,
  \item population model,
  \item parent selection mechanism,
  \item variation operators (mutation and crossover), and
  \item initialization procedure and termination condition.
\end{itemize}
The evaluation function depends on a given problem. The self-adaptive evolutionary algorithm uses the population model $(\mu,\lambda)$, where the $\lambda$ offsprings is generated from the $\mu$ parents. However, the parents that are selected with tournament selection \citep{book:eiben2003} are replaced by the $\mu$ the best offsprings according to the appropriate population model. The ratio $\lambda /\mu \approx 7$ is used for the efficient self-adaptation \citep{book:eiben2003}. Typically, the normal uncorrelated mutation with $n$ step sizes, discrete and arithmetic crossover are used by the HSA-EA. Normally, the probabilities of mutation and crossover are set according to the given problem. Selection of the suitable local search heuristics that improve the current solution is a crucial for the performance of the HSA-EA. On the other hand, the implementation of neutral survivor selection is straightforward. Finally, the scheme of the HSA-EA is represented in the Algorithm~\ref{alg:HSA-EA}.

\begin{algorithm}[t!]
\caption{Hybrid Self-Adaptive Evolutionary Algorithm.}
\label{alg:HSA-EA}
\begin{algorithmic}[1]
\STATE $t=0$;
\STATE $Q^{(0)}= initialization\_procedure()$;
\STATE $P^{(0)}= evaluate\_and\_improve(Q^{(0)})$;
\WHILE { {\bf not} termination\_condition }
    \STATE $P' = select\_parent(P^{(t)})$;
    \STATE $P'' = mutate\_and\_crossover(P')$;
    \STATE $P''' = evaluate\_and\_improve(P'')$;
    \STATE $P^{(t+1)}=select\_survivor(P''')$;
    \STATE $t=t+1$;
\ENDWHILE
\end{algorithmic}
\end{algorithm}

In the rest of the chapter we present the implementation of the HSA-EA for the graph 3-coloring. This algorithm is hybridized with the DSatur \citep{art:brelaz1979} construction heuristic that is well-known traditional heuristic for the graph 3-coloring.

\subsection{Graph 3-coloring}

Graph 3-coloring can be informally defined as follows. Let assume, an undirected graph $G=(V,E)$ is given, where $V$ denotes a finite set of vertices and $E$ a finite set of unordered pairs of vertices named edges \citep{book:murty2008}. The vertices of graph $G$ have to be colored with three colors such that no one of vertices connected with an edge is not colored with the same color.

Graph 3-coloring can be formalized as constraint satisfaction problem (CSP) that is denoted as a pair $\langle S,\phi \rangle$, where $S$ denotes a free search space and $\phi$ a Boolean function on $S$. The free search space denotes the domain of candidate solutions $x\in S$ and does not contain any constraints, i.e., each candidate solution is feasible. The function $\phi$ divides the search space $S$ into feasible and unfeasible regions. The solution of constraint satisfaction problem is found when all constraints are satisfied, i.e., when $\phi (x) = true$.

However, for the 3-coloring of graph $G=(V,E)$ the free search space $S$ consists of all permutations of vertices $v_{i} \in V$ for $i=1 \ldots n$. On the other hand, the function $\phi$ (also feasibility condition) is composed of constraints on vertices. That is, for each vertex $v_{i} \in V$ the corresponding constraint $C^{v_{i}}$ is defined as the set of constraints involving vertex $v_{i}$, i.e., edges $(v_{i},v_{j})\in E$ for $j=1\ldots m$ connecting to vertex $v_{i}$. The feasibility condition is expressed as conjunction of all constraints $\phi (x)=\wedge _{v_{i} \in V} C^{v_{i}}(x)$.

Handling direct constraints in evolutionary algorithms is not straightforward. To overcome this problem, the constraint satisfaction problems are, typically, transformed into unconstrained (also free optimization problem) by the sense of a penalty function. The more an infeasible solution is far away from feasible region, the higher is the penalty. Moreover, this penalty function can act as an evaluation function by the evolutionary algorithm. For graph 3-coloring it can be expressed as

\begin{equation}\label{eq:fit}
 f(x)=\sum_{i=0}^{n} \psi(x,C^{v_{i}}),
\end{equation}

where the function $\psi(x,C^{v_{i}})$ is defined as

\begin{equation}\label{eq:viol}
\psi(x,C^{v_{i}})=\left\{\begin{matrix}
1 & \textup{if}\ x\ \textup{violates\ at\ least\ one\ } c_{j}\in C^{v_{i}}, \\
0 & \textup{otherwise}.
\end{matrix}\right.
\end{equation}

Note that all constraints in solution $x\in S$ are satisfied, i.e., $\phi (x)=true$ if and only if $f(x)=0$. In this way, the Equation~\ref{eq:fit} represents the feasibility condition and can be used to estimate the quality of solution $x\in S$ in the permutation search space. The permutation $x$ determines the order in which the vertices need to be colored. The size of the search space is huge, i.e., $n!$. As can be seen from Equation~\ref{eq:fit}, the evaluation function depends on the number of constraint violations, i.e., the number of uncolored vertices. This fact causes that more solutions can have the same value of the evaluation function. Consequently, the large neutral networks can arise \citep{inbook:stadler1995}. However, the neutral solutions are avoided if the slightly modified evaluation function is applied, as follows:

\begin{equation}\label{eq:fit1}
 f(x)=\sum_{i=0}^{n} w_{i}\times \psi(x,C^{v_{i}}),\ \ \ \ w_{i} \neq 0,
\end{equation}

where $w_{i}$ represents the weight. Higher than the value of weights harder the appropriate vertex is to color.

\subsubsection{The Hybrid Self-adaptive Evolutionary Algorithm for Graph 3-coloring}

The hybrid self-adaptive evolutionary algorithm is hybridized with the DSatur \citep{art:brelaz1979} construction heuristic and the local search heuristics. In addition, the problem specific knowledge is incorporated by the initialization procedure and the neutral survivor selection. In this section we concentrate, especially, on a description of those elements in evolutionary algorithm that incorporate the problem specific knowledge. That are:
\begin{itemize}
  \item the initialization procedure,
  \item the genotype-phenotype mapping,
  \item local search heuristics and
  \item the neutral survivor selection.
\end{itemize}
The other elements of this evolutionary algorithm, as well as neutral survivor selection, are common and therefore, discussed earlier in the chapter.
\\\\
\textit{The Initialization Procedure}
\\
Initially, original DSatur algorithm orders the vertices $v_{i}\in V$ for $i=1 \ldots n$ of a given graph $G$ descendingly according to the vertex degrees denoted by $d_{G}(v_{i})$ that counts the number of edges that are incident with the vertex $v_{i}$ \citep{book:murty2008}. To simulate behavior of the original DSatur algorithm \citep{art:brelaz1979}, the first solution in the population is initialized as follows:
\begin{equation}\label{eq:init}
 y_{i}^{(0)}=\frac{ d_{G}(v_{i})}{\textup{max}_{i=1\ldots n}d_{G}(v_{i})}, \ \ \ \ \textup{for}\ i=1 \ldots n.
\end{equation}
Because the genotype representation is mapped into a permutation of weights by decoder the same ordering as by original DSatur is obtained, where the solution can be found in the first step. However, the other $\mu-1¢$ solutions in the population are initialized randomly.
\\\\
\textit{The Genotype-phenotype mapping}
\\
As illustrated in Fig.~\ref{pic:3}, the solution is represented in genotype search space as tuple $\langle y_{1}, \ldots ,y_{n}, \sigma_{1}, \ldots , \sigma_{n} \rangle$, where problem variables $y_{i}$ for $i=1 \ldots n$ denote how hard the given vertex is to color and control parameters $ \sigma_{i}$ for $i=1 \ldots n$ mutation steps of normal mutation. A decoder decodes the problem variables into permutation of vertices and corresponding weights. However, all feasible permutation of vertices form the permutation search space. The solution in this search space is represented as tuple $\langle v_{1}, \ldots ,v_{n},w_{1}, \ldots ,w_{n} \rangle$, where variables $v_{i}$ for $i=1 \ldots n$ denote the permutation of vertices and variables $w_{i}$ corresponding weights. The vertices are ordered into permutation so that vertex $v_{i}$ is predecessor of vertex $v_{i+1}$ if and only if $w_{i} \geq w_{i+1}$. Values of weights $w_{i}$ are obtained by assigning the corresponding values of problem variables, i.e. $w_{i}=y_{i}$ for $i=1 \ldots n$. Finally, DSatur construction heuristic maps the permutation of vertices and corresponding weights into phenotypic solution space that consists of all possible 3-colorings $c_{i}$. Note that the size of this space is $3^{n}$. DSatur construction heuristic acts like original DSatur algorithm \citep{art:brelaz1979}, i.e. it takes the permutation of vertices and color these as follows:
\begin{itemize}
  \item the heuristic selects a vertex with the highest saturation, and colors it with the lowest of the three colors;
  \item in the case of a tie, the heuristic selects a vertex with the maximal weight;
  \item in the case of a tie, the heuristic selects a vertex randomly.
\end{itemize}
The main difference between this heuristic and the original DSatur algorithm is in the second step where the heuristic selects the vertices according to the weights instead of degrees.
\\\\
\textit{Local Search Heuristics}
\\
The current solution is improved by a sense of local search heuristics. At each evaluation of solution the best neighbor is obtained by acting of the following original local search heuristics:
\begin{itemize}
  \item inverse,
  \item ordering by saturation,
  \item ordering by weights, and
  \item swap.
\end{itemize}

The evaluation of solution is presented in Algorithm~\ref{alg:eval} from which it can be seen that the local search procedure ($k\_move(y)$) is iterated until improvements are perceived. However, this procedure implements all four mentioned local search heuristics. The best neighbor is generated from the current solution by local search heuristics with $k$-exchanging of vertices. In the case, the best neighbor is better than the current solution the later is replaced by the former.

\begin{algorithm}[t!]
\caption{Evaluate and improve. $y$: solution.}
\label{alg:eval}
\begin{algorithmic}[1]
\STATE $est=evaluate(y)$;
\REPEAT
    \STATE $climbing=FALSE$;
    \STATE $y' = k\_move(y)$;
    \STATE $ls\_est = evaluate(y')$;
    \IF {$ls\_est < est$}
        \STATE $y=y'$;
        \STATE $est=ls\_est$;
        \STATE $climbing=TRUE$;
    \ENDIF
\UNTIL {$climbing=TRUE$}
\end{algorithmic}
\end{algorithm}

\begin{figure}[htb]	
\centering
\includegraphics[width=10cm]{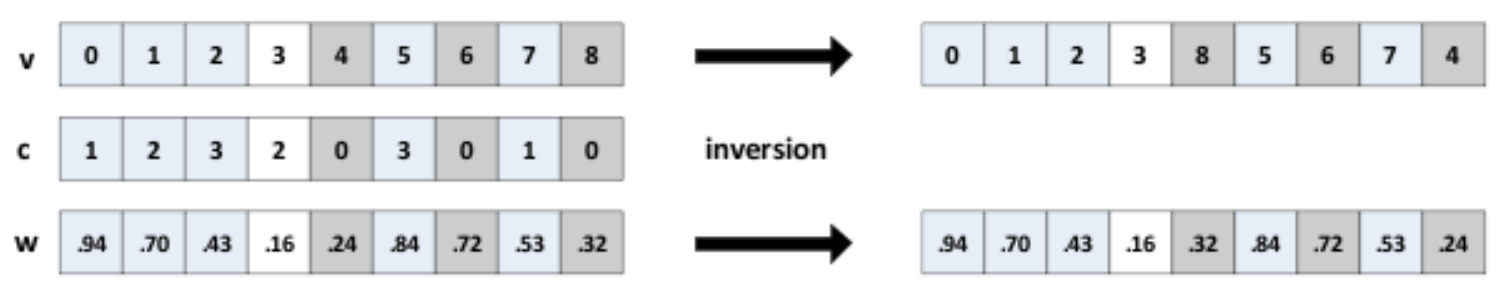} 
\caption{Inverse local search heuristic}
\label{pic:inverse}
\end{figure}

\begin{figure}[htb]	
\centering
\includegraphics[width=12cm]{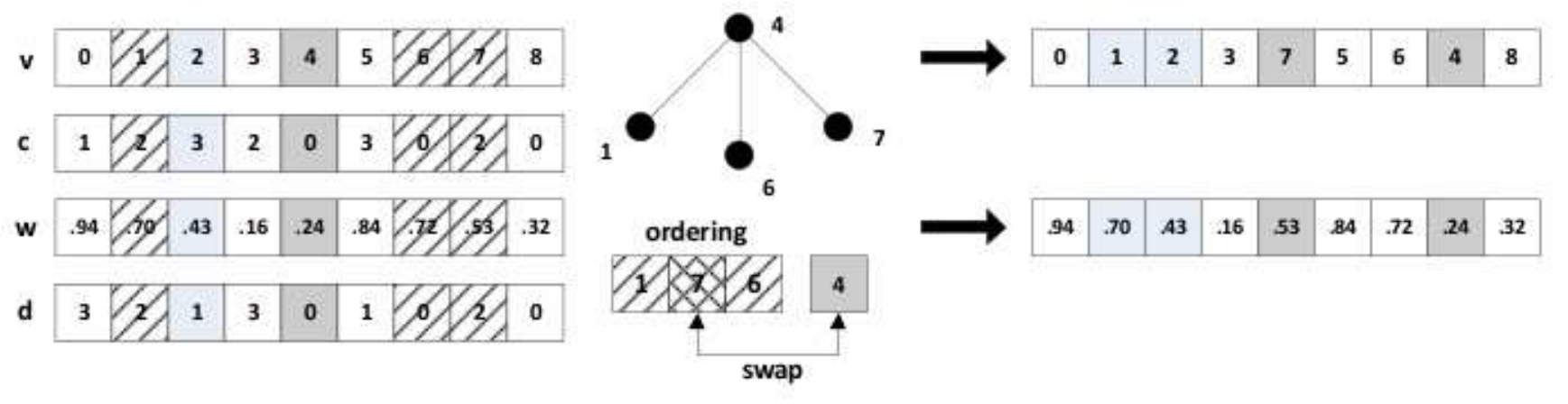} 
\caption{Ordering by saturation local search heuristic}
\label{pic:order}
\end{figure}

\begin{figure}[htb]	
\centering
\includegraphics[width=12cm]{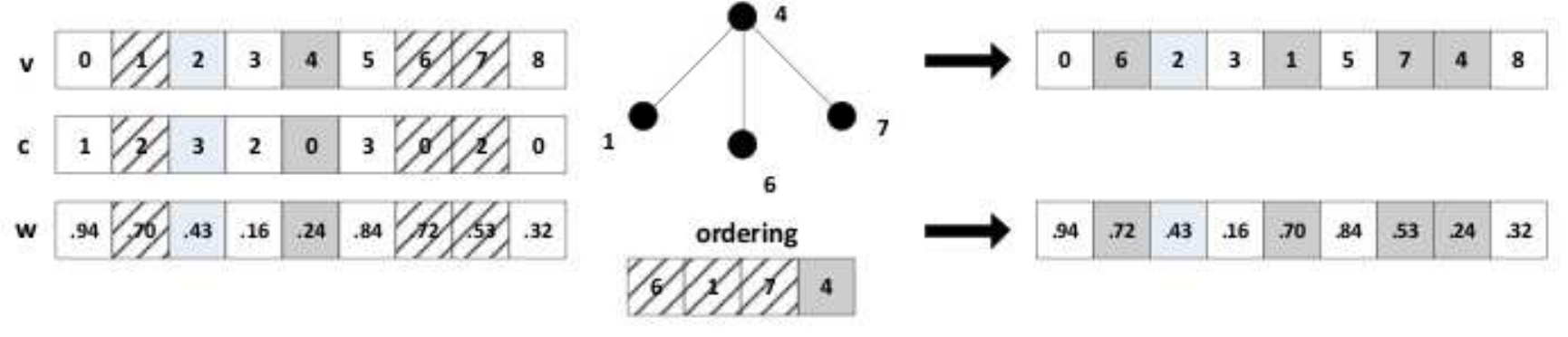} 
\caption{Ordering by weights local search heuristic}
\label{pic:order2}
\end{figure}

\begin{figure}[htb]	
\centering
\includegraphics[width=10cm]{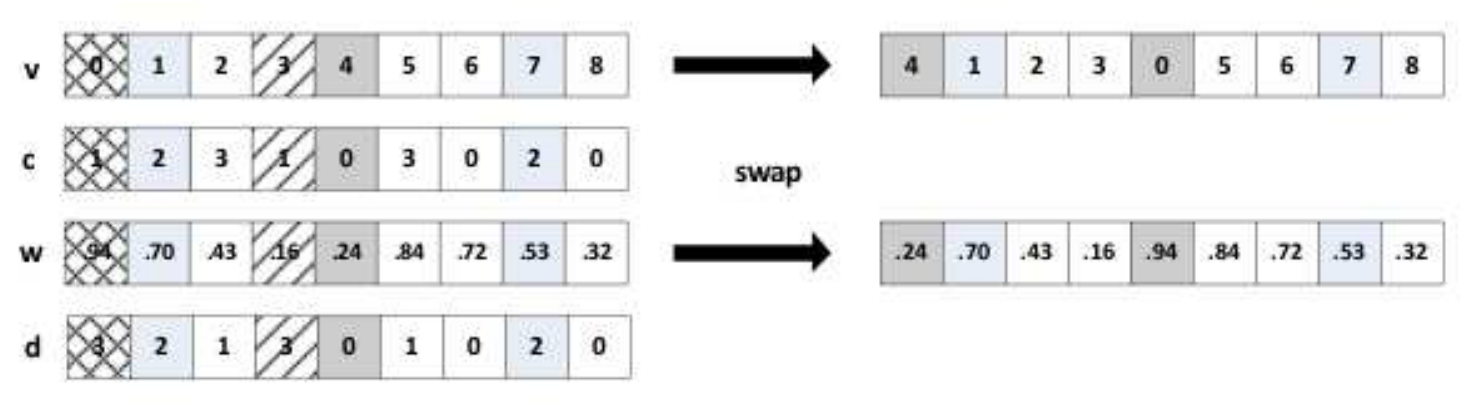}
\caption{Swap local search heuristic}
\label{pic:swap}
\end{figure}

In the rest of the subsection, an operation of the local search heuristics is illustrated in Fig.~\ref{pic:inverse}-\ref{pic:swap} by samples, where a graph with nine vertices is presented. The graph is composed of a permutation of vertices $v$, corresponding coloring $c$, weights $w$ and saturation degrees $d$.


The inverse local search heuristic finds all uncolored vertices in a solution and inverts their order. As can be shown in Fig.~\ref{pic:inverse}, the uncolored vertices 4, 6 and 8 are shadowed. The best neighbor is obtained by inverting of their order as is presented on right-hand side of this figure. The number of vertex exchanged is dependent of the number of uncolored vertices ($k \textendash opt$ neighborhood).

The ordering by saturation local search heuristic acts as follows. The first uncolored vertex is taken at the first. To this vertex a set of adjacent vertices are selected. Then, these vertices are ordered descending with regard to the values of saturation degree. Finally, the adjacent vertex with the highest value of saturation degree in the set of adjacent vertices is swapped with the uncolored vertex. Here, the simple $1 \textendash opt$ neighborhood of current solution is defined by this local search heuristic. In the example on Fig.~\ref{pic:order} the first uncolored vertex 4 is shadowed, while its adjacent vertices 1, 6 and 7 are hatched. However, the vertices 1 and 7 have the same saturation degree, therefore, the vertex 7 is selected randomly. Finally, the vertices 4 in 7 are swapped (right-hand side of Fig.~\ref{pic:order}).

When ordering of weights, the local search heuristic takes the first uncolored vertex and determines a set of adjacent vertices including it. This set of vertices is then ordered descending with regard to the values of weights. This local search heuristic determines the $k \textendash opt$ neighborhood of current solution, where $k$ is dependent of a degree of the first uncolored vertex. As illustrated by Fig.~\ref{pic:order2}, the uncolored vertex 4 is shadowed, while its adjacent vertices 1, 6 and 7 are hatched. The appropriate ordering of the selected set of vertices is shown in the right-hand of Fig.~\ref{pic:order2} after the operation of the local search heuristic.

The swap local search heuristic finds the first uncolored vertex and descendingly orders the set of all predecessors in the solution according to the saturation degree. Then, the uncolored vertex is swapped with the vertex from the set of predecessors with the highest saturation degree. When more vertices with the same highest saturation degree are arisen, the subset of these vertices is determined. The vertex from this subset is then selected randomly. Therefore, the best neighbor of the current solution is determined by an exchange of two vertices ($1 \textendash opt$ neighborhood). As illustrated in Fig.~\ref{pic:swap}, the first uncolored vertex 4 is shadowed, while the vertices 0 and 4 that represent the subset of vertices with the highest saturation are hatched. In fact, the vertex 0 is selected randomly and the vertices 0 and 4 are swapped as is presented in right-hand of Fig.~\ref{pic:swap}.

\subsubsection{Analysis of the Hybrid Self-adaptive Evolutionary Algorithm for Graph 3-coloring}
The goal of this subsection is twofold. At the first, an influence of the local search heuristics on results of the HSA-EA is analyzed in details. Further, a comparison of the HSA-EA hybridized with the neutral survivor selection and the HSA-EA with the deterministic selection is made. In this context, the impact of the heuristic initialization procedure are taken into consideration as well.

Characteristics of the HSA-EA used in experiments were as follows. The normal distributed mutation was employed and applied with mutation probability of 1.0. The crossover was not used. The tournament selection with size 3 selects the parents for mutation. The population model $(15,100)$ was suitable for the self-adaptation because the ratio between parents and generated offspring amounted to $100/15 \approx 7$ as recommended by \cite{book:baeck1996}. As termination condition, the maximum number of evaluations to solution was used. Fortunately, the average number of evaluations to solution ($AES$) that counts the number of evaluation function calls was employed as the performance measure of efficiency. In addition, the average number of uncolored nodes ($AUN$) was employed as the performance measure of solution quality. This measure was applied when the HSA-EA does not find the solution and counts the number of uncolored vertices. Nevertheless, the success rate ($SR$) was defined as the primary performance measure and expressed as the ratio between the runs in which the solution was found and all performed runs.

The \cite{website:Culberson2008} random graph generator was employed for generation of random graphs that constituted the test suite. It is capable to generate the graphs of various types, number of vertices, edge densities and seeds of random generator. In this study we concentrated on the equi-partite type of graphs. This type of graphs is not the most difficult to color but difficult enough for many existing algorithms \citep{inbook:culberson1996}. The random graph generator divides the vertices of graph into three color sets before generating randomly. In sense of equi-partite random graph, these color sets are as close in size as possible.

All generated graphs consisted of $n=1,000$ vertices. An edge density is controlled by parameter $p$ of the random graph generator that determines probability that two vertices $v_{i}$ and $v_{j}$ in the graph $G$ are connected with an edge $(v_{i},v_{j})$ \citep{inproc:chiarandini2010}. However, if $p$ is small the graph is not connected because the edges are sparse. When $p$ is increased the number of edges raised and the graph becomes interconnected. As a result, the number of constraints that needs to be satisfied by the coloring algorithm increases until suddenly the graph becomes uncolorable. This occurrence depends on a ratio between the number of edges and the number of vertices. The ratio is referred to as the threshold \citep{art:hayes2003}. That is, in the vicinity of the threshold the vertices of the random generated graph becomes hard to color or even the graph becomes uncolorable. Fortunately, the graph instances with this ratio much higher that the threshold are easy to color because these graphs are densely interconnected. Therefore, many global optima exist in the search space that can be discovered easy by many graph 3-coloring algorithms. Interestingly, for random generated graphs the threshold arises near to the value 2.35 \citep{art:hayes2003}. For example, the equi-partite graph generated with number of vertices $1,000$ and the edge density determined by $p=0.007$ consists of 2,366 edges. Because the ratio $2,366/1,000=2.37$ is near to the threshold, we can suppose that this instance of graph is hard to color. The seed $s$ of random graph generator determines which of the two vertices $v_{i}$ and $v_{j}$ are randomly drawn from different 3-color sets to form an edge $(v_{i},v_{j})$ but it does not affect the performance of the graph 3-coloring algorithm \citep{art:eiben1998}. In this study, the instances of random graphs with seed $s=5$ were employed.

\begin{figure}[H]	
\centering
\subfigure[Success rate (SR)] {\includegraphics[width=6.4cm]{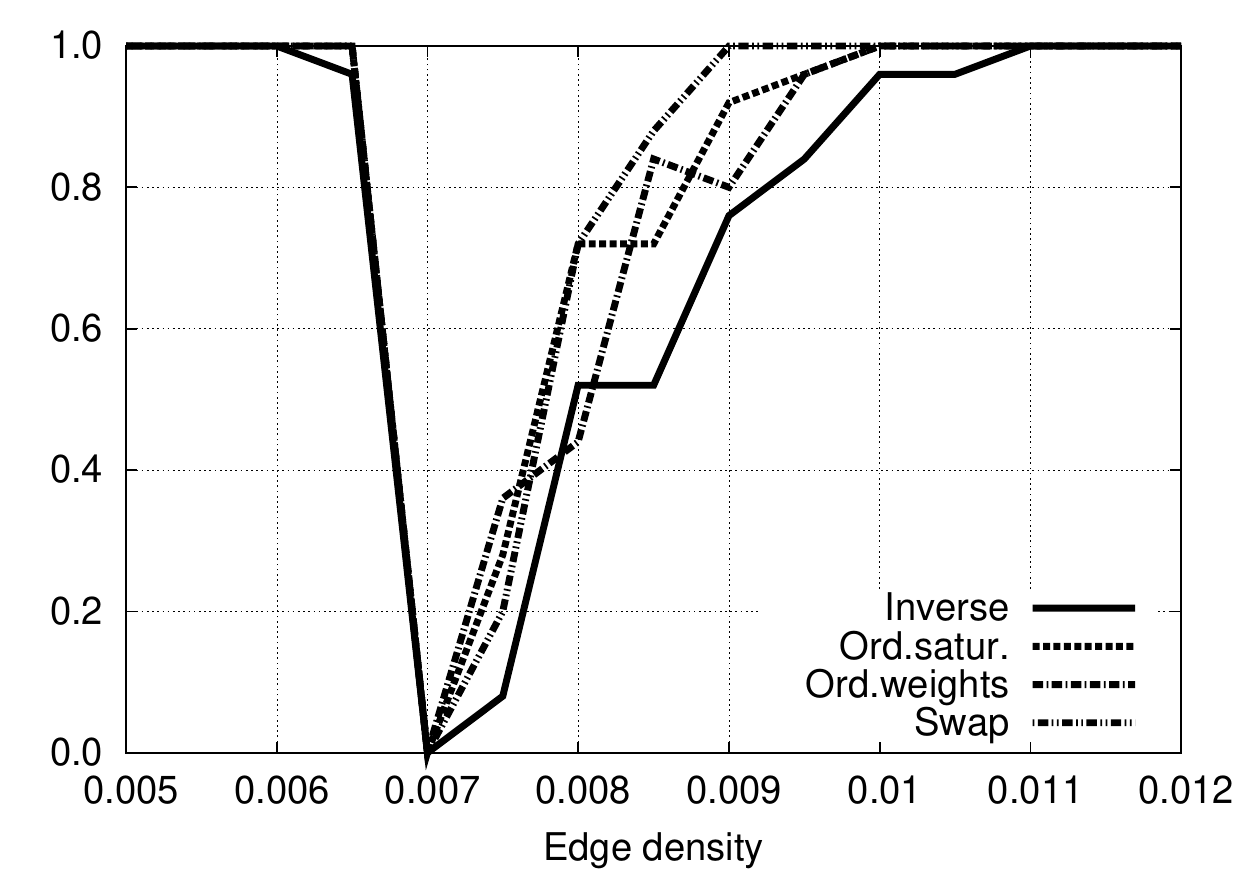}}
\subfigure[Success rate (SR)] {\includegraphics[width=6.4cm]{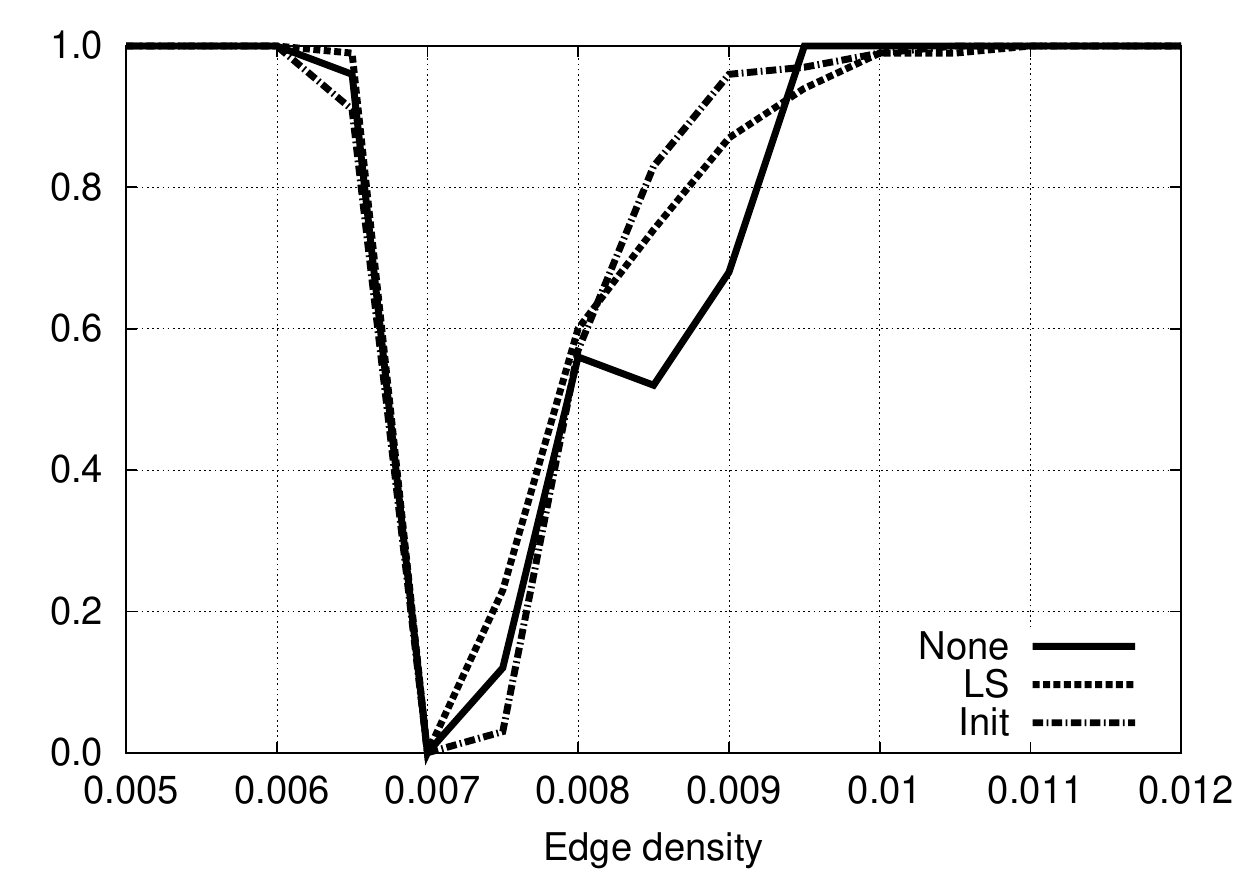}}
\subfigure[Average evaluations to solution (AES)] {\includegraphics[width=6.4cm]{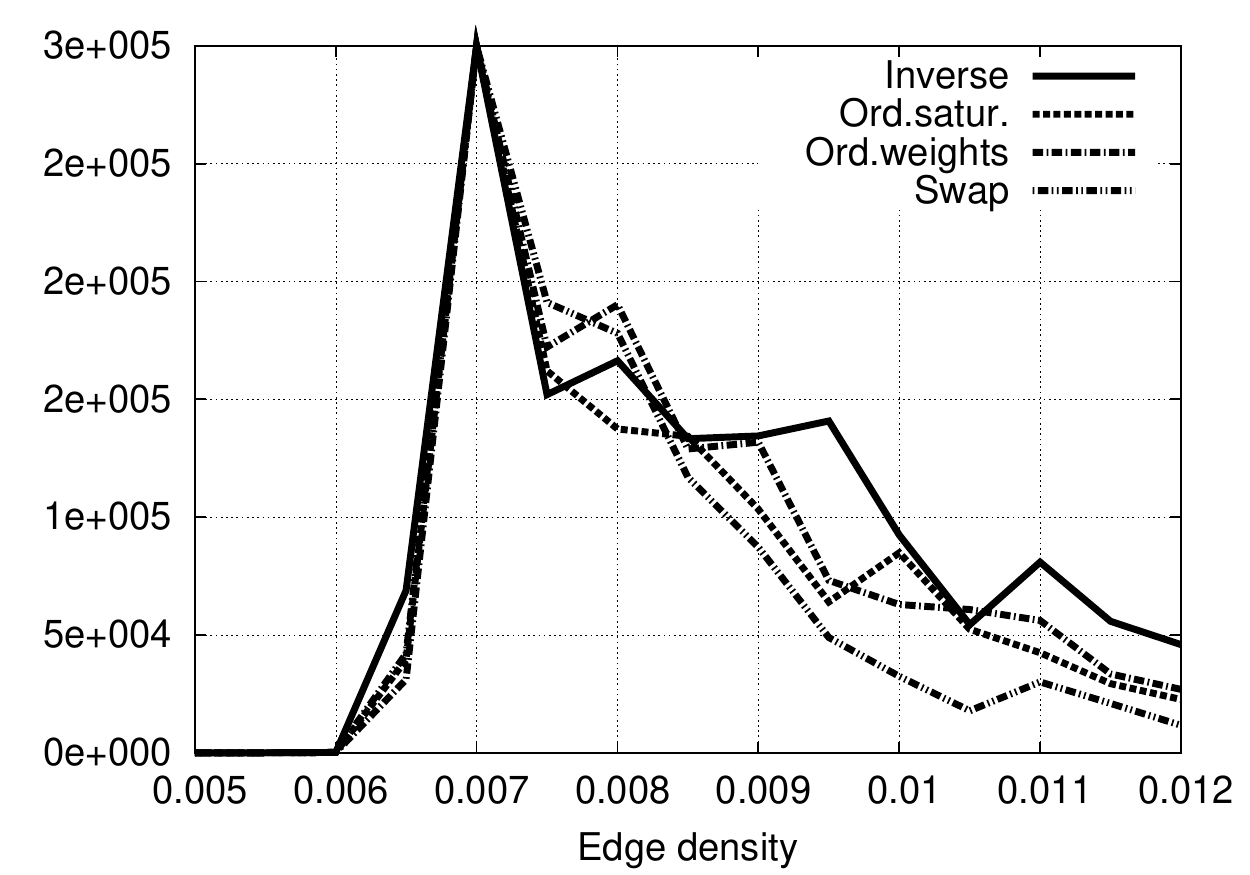}}
\subfigure[Average evaluations to solution (AES)] {\includegraphics[width=6.4cm]{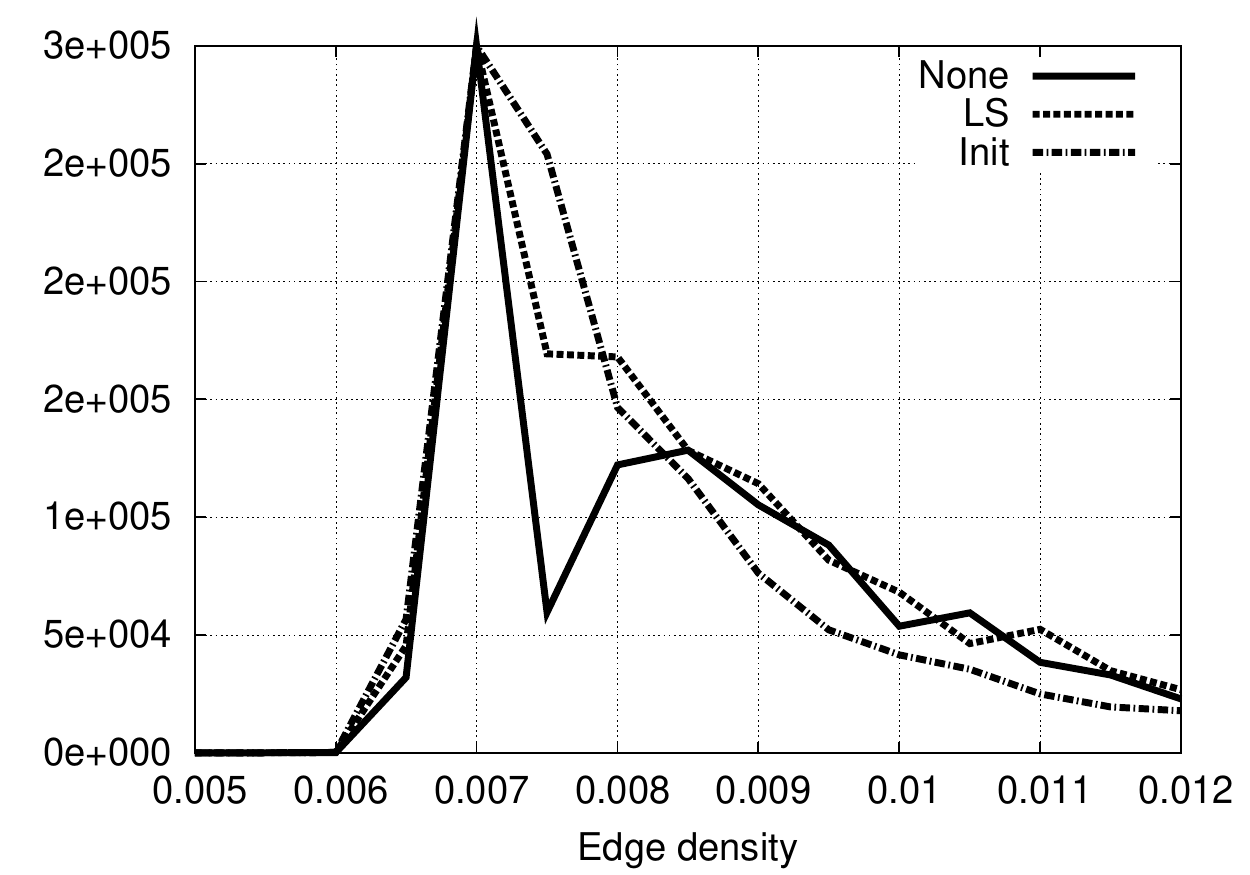}}
\subfigure[Average number of uncolored nodes (AUN)] {\includegraphics[width=6.4cm]{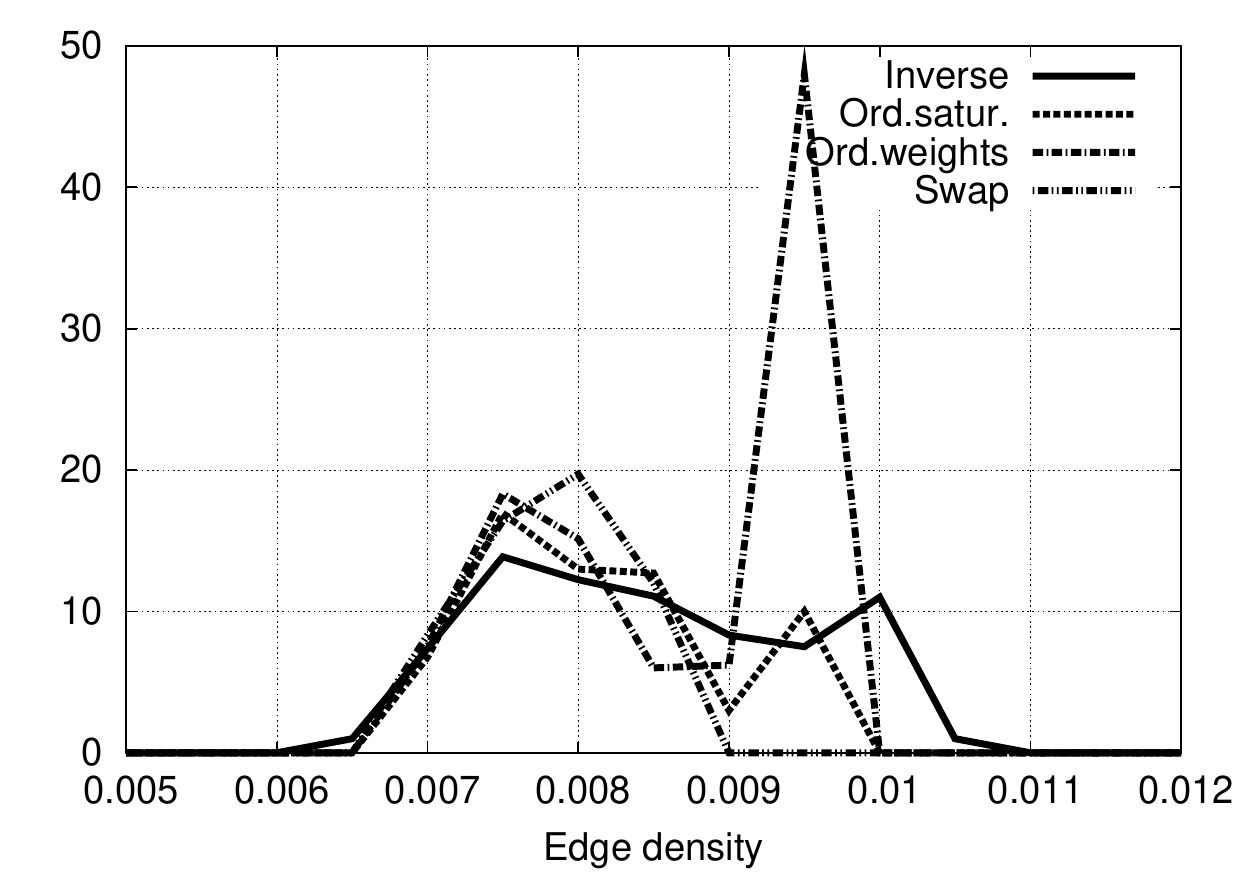}}
\subfigure[Average number of uncolored nodes (AUN)] {\includegraphics[width=6.4cm]{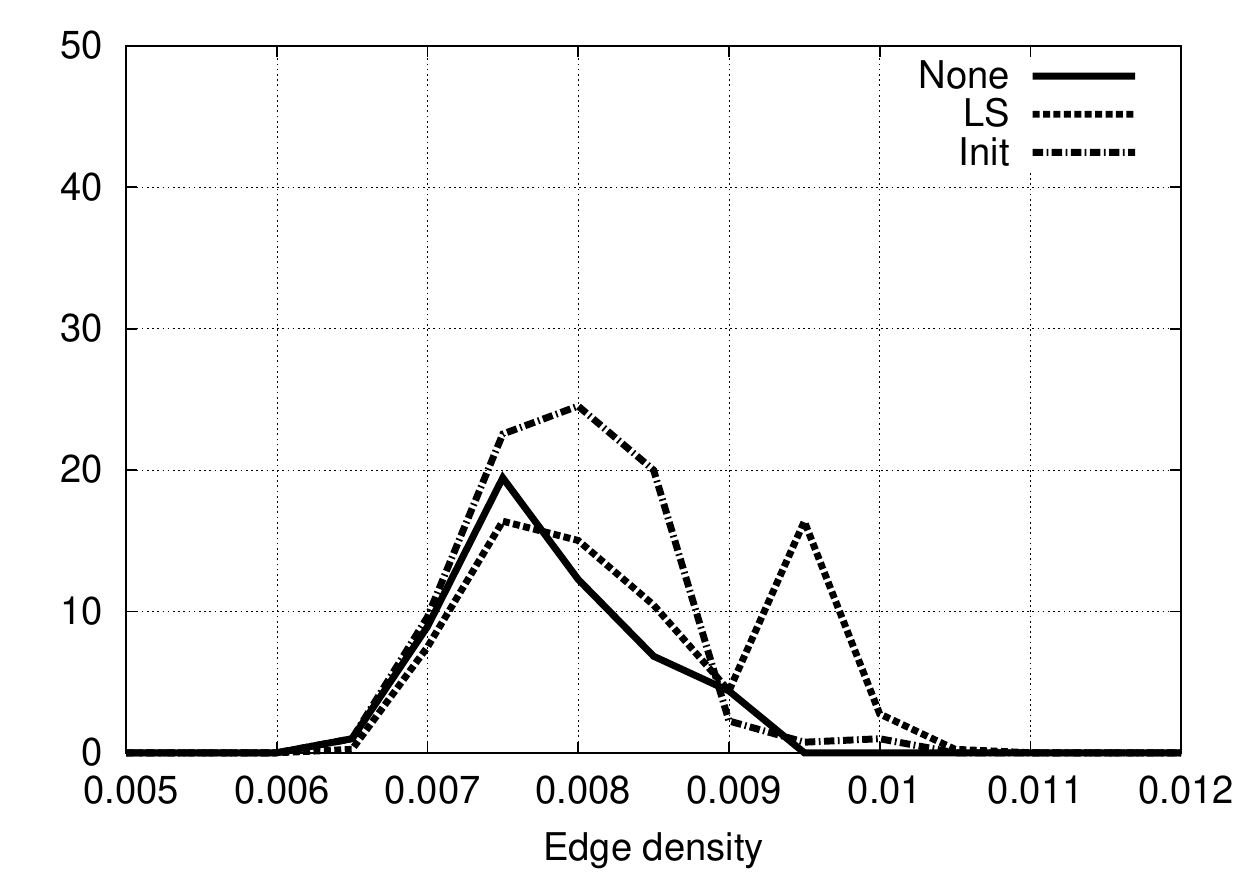}}
\caption{Influence of local search heuristics on results of HSA-EA solving equi-partite graphs}
\label{fig:Sub_3}
\end{figure}

To capture a phenomenon of the threshold, the parameter $p$ by generation of the equi-partite graphs was varied from $p=0.005$ to $p=0.012$ in a step of $0.0005$. In this way, the test suite consisted of 15 instances of graphs, in which the hardest graph with $p=0.007$ was presented as well. In fact, the evolutionary algorithm was applied to each instance $25$ times and the average results of these runs were considered.
\\\\
\textit{The impact of the local search heuristics}
\\
In this experiments, the impact of four implemented local search heuristics on results of the HSA-EA was taken into consideration. Results of the experiments are illustrated in the Fig.~\ref{fig:Sub_3} that is divided into six graphs and arranged according to the particular measures $SR$, $AES$ and $AUN$. The graphs on the left side of the figure, i.e. 8.a, 8.c and 8.e, represent a behavior of the HSA-EA hybridized with four different local search heuristics. This kind of the HSA-EA is referred to as original HSA-EA in the rest of chapter.

A seen by the Fig.~\ref{fig:Sub_3}.a, no one of the HSA-EA versions was succeed to solve the hardest instance of graph with $p=0.007$. The best results in the vicinity of the threshold is observed by the HSA-EA hybridizing with the ordering by saturation local search heuristic ($SR=0.36$ by $p=0.0075$). The overall best performance is shown by the HSA-EA using the swap local search heuristic. Although the results of this algorithm is not the best by instances the nearest to the threshold ($SR=0.2$ by $p=0.0075$), this local search heuristic outperforms the other by solving the remaining instances in the collection.

In average, results according to the $AES$ (Fig.~\ref{fig:Sub_3}.c) show that the HSA-EA hybridized with the swap local search heuristic finds the solutions with the smallest number of the fitness evaluations. However, troubles are arisen in the vicinity of the threshold, where the HSA-EA with other local search heuristics are faced with the difficulties as well. Moreover, at the threshold the HSA-EA hybridizing with all the used local search heuristics reaches the limit of 300,000 allowed function evaluations.

The HSA-EA hybridizing with the ordering by saturation local search heuristic demonstrates the worst results according to the $AUN$, as presented in the Fig.~\ref{fig:Sub_3}.e. The graph instance by $p=0.0095$ was exposed as the most critical by this algorithm ($AUN=50$) although this is not the closest to the threshold. In average, when the HSA-EA was hybridized with the other local search heuristics than the ordering by saturation, all instances in the collection were solved with less than 20 uncolored vertices.

In the right side of the Fig.~\ref{fig:Sub_3}, results of different versions of the HSA-EA are collected. The first version that is designated as $None$ operates with the same parameters as the original HSA-EA but without the local search heuristics. The label $LS$ in this figure indicates the original version of the HSA-EA. Finally, the label $Init$ denotes the original version of the HSA-EA with the exception of initialization procedure. While all considered versions of the HSA-EA uses the heuristic initialization procedure this version of the algorithm employs the pure random initialization. Note, in the figure, results for this version of the HSA-EA were obtained after 25 runs, while for the versions of the HSA-EA with the local search heuristics the average results were obtained after performing of all four local search heuristics, i.e. after $100$ runs.

In Fig.~\ref{fig:Sub_3}.b results of different versions of the HSA-EA according to the $SR$ are presented. The best results by the instances the nearest to the threshold ($p \in [0.0075 \ldots 0.008]$) are observed by the original HSA-EA. Conversely, the HSA-EA with the random initialization procedure ($Init$) gained the worst results by the instances the nearest to the threshold, while these were better while the edge density was raised regarding the original HSA-EA. The turning point represents the instance of graph with $p=0.008$. After this point is reached the best results were overtaken by the HSA-EA with the random initialization procedure ($Init$).

In contrary, the best results by the instances the nearest to the threshold according to the $AES$ was observed by the HSA-EA without local search heuristics ($None$). Here, the turning point regarding the performance of the HSA-EA ($p=0.008$) was observed as well. After this point results of the HSA-EA without local search heuristics becomes worse. Conversely, the HSA-EA with random initialization procedure that was the worst by the instances before the turning point becomes the best after this.

\begin{figure}[htb]	 %
\centering
\subfigure[Number of uncolored nodes by neutral selection] {\includegraphics[width=6.4cm]{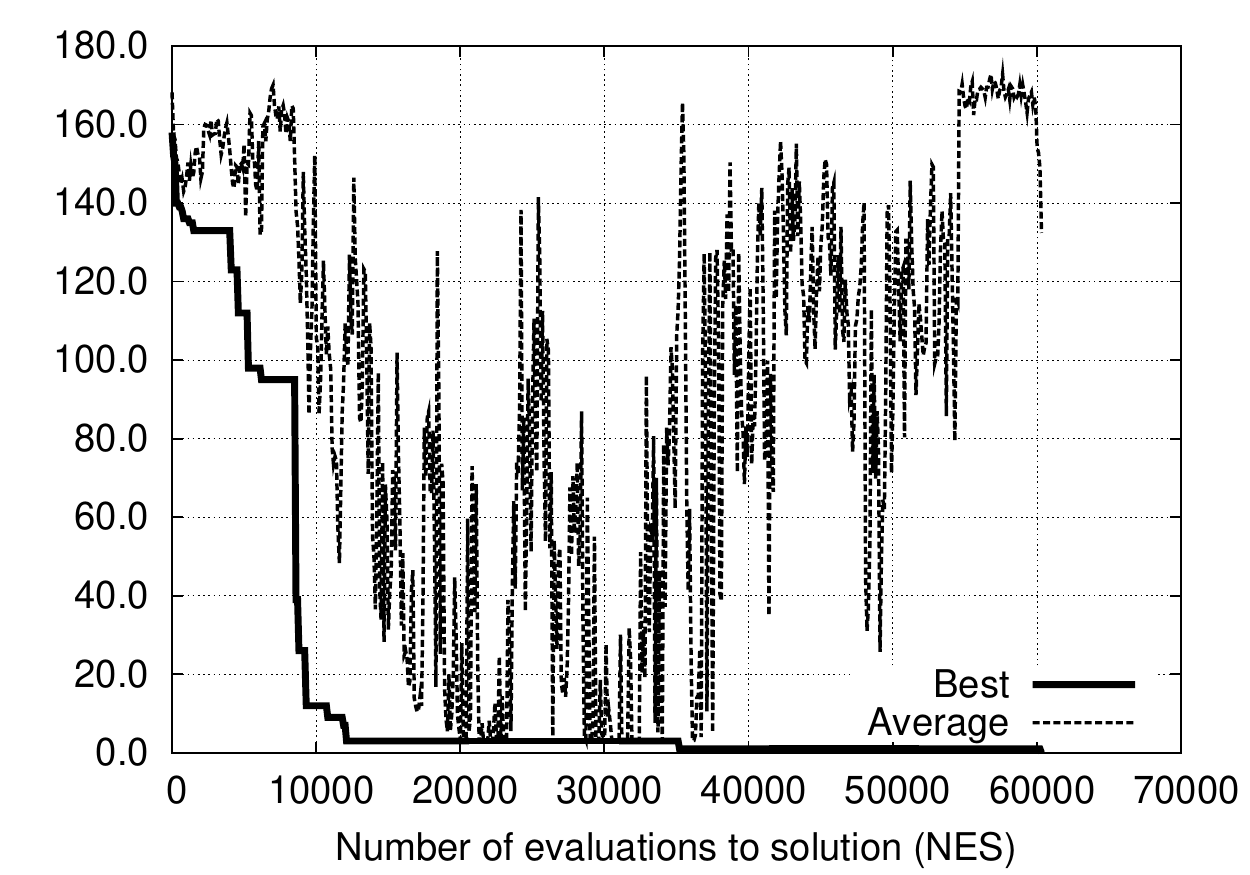}}
\subfigure[Percent of neutral solutions] {\includegraphics[width=6.4cm]{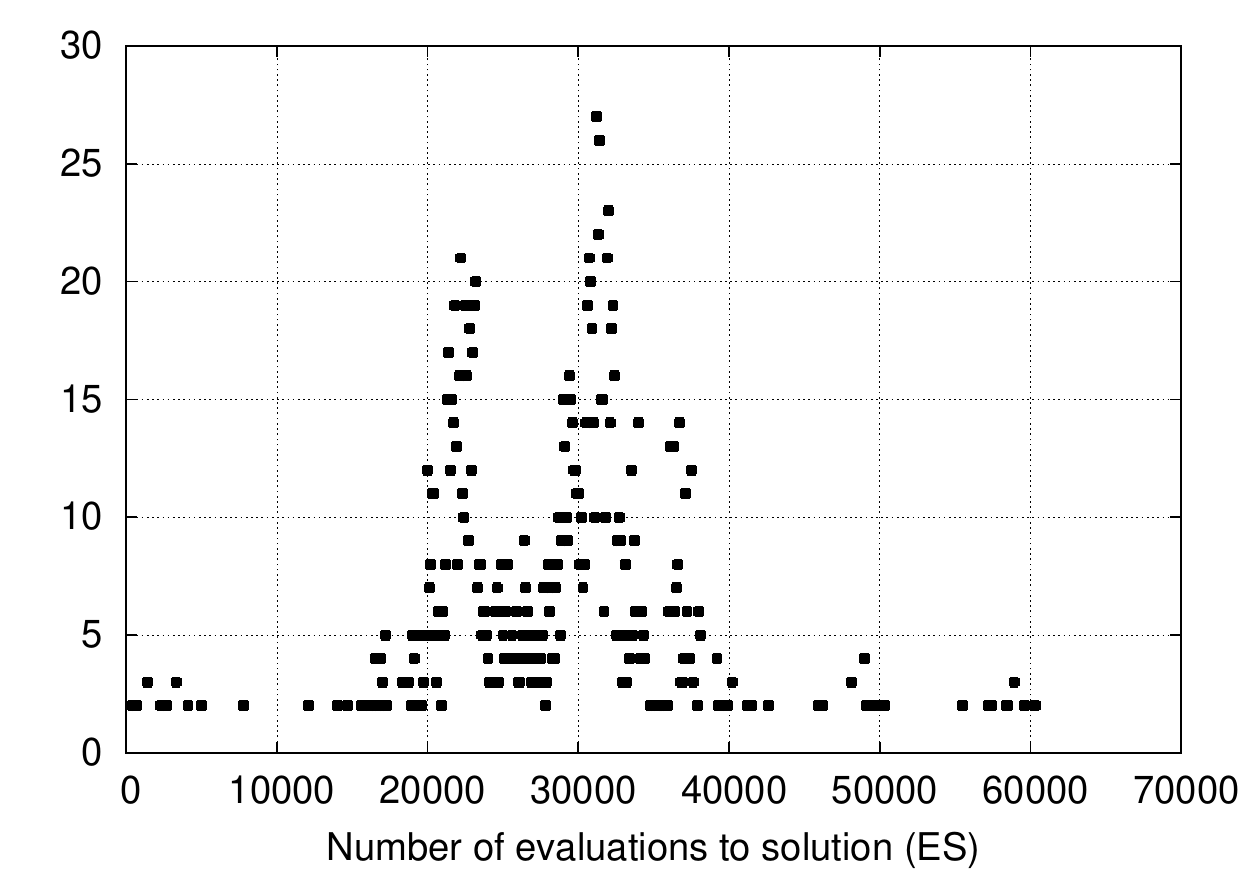}}
\subfigure[Number of uncolored nodes by det. selection] {\includegraphics[width=6.4cm]{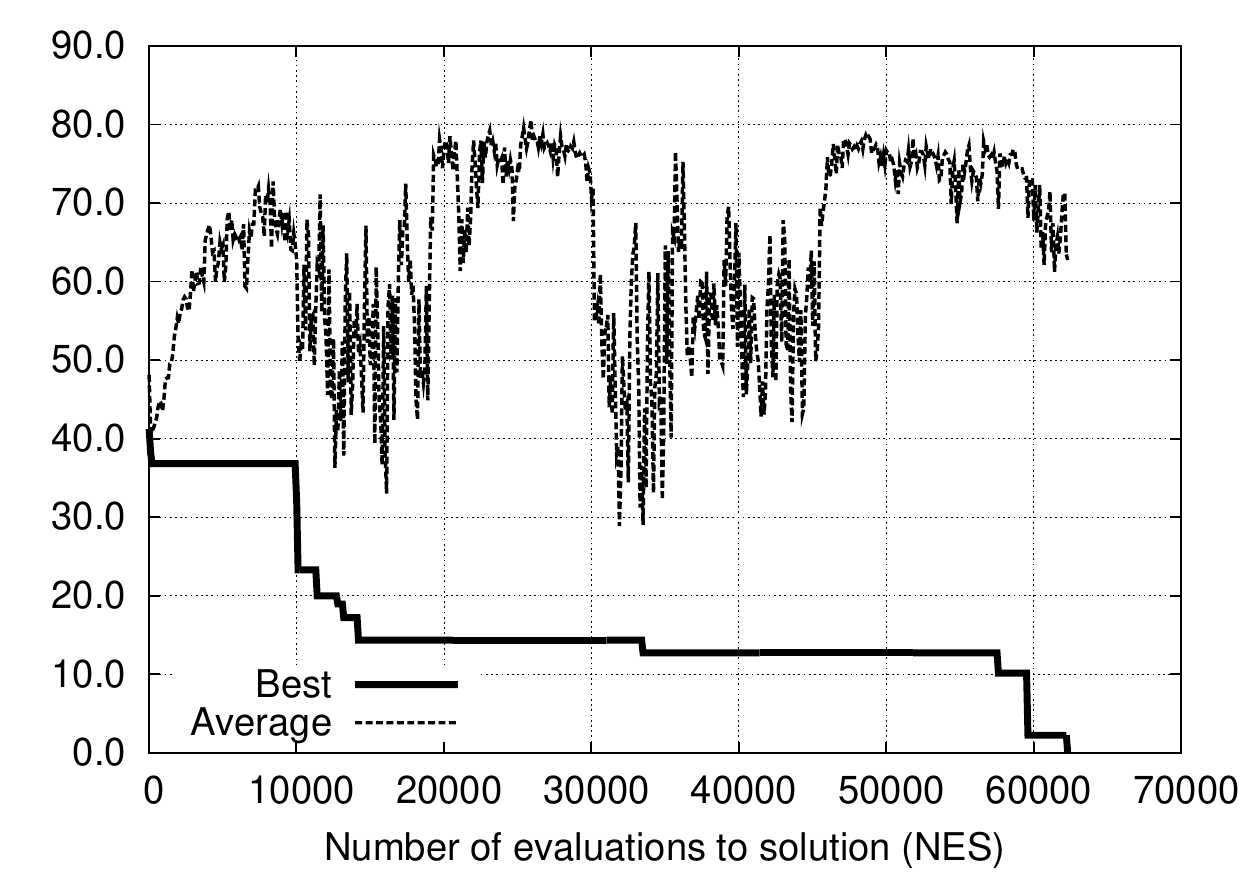}}
\subfigure[Diversity of population] {\includegraphics[width=6.4cm]{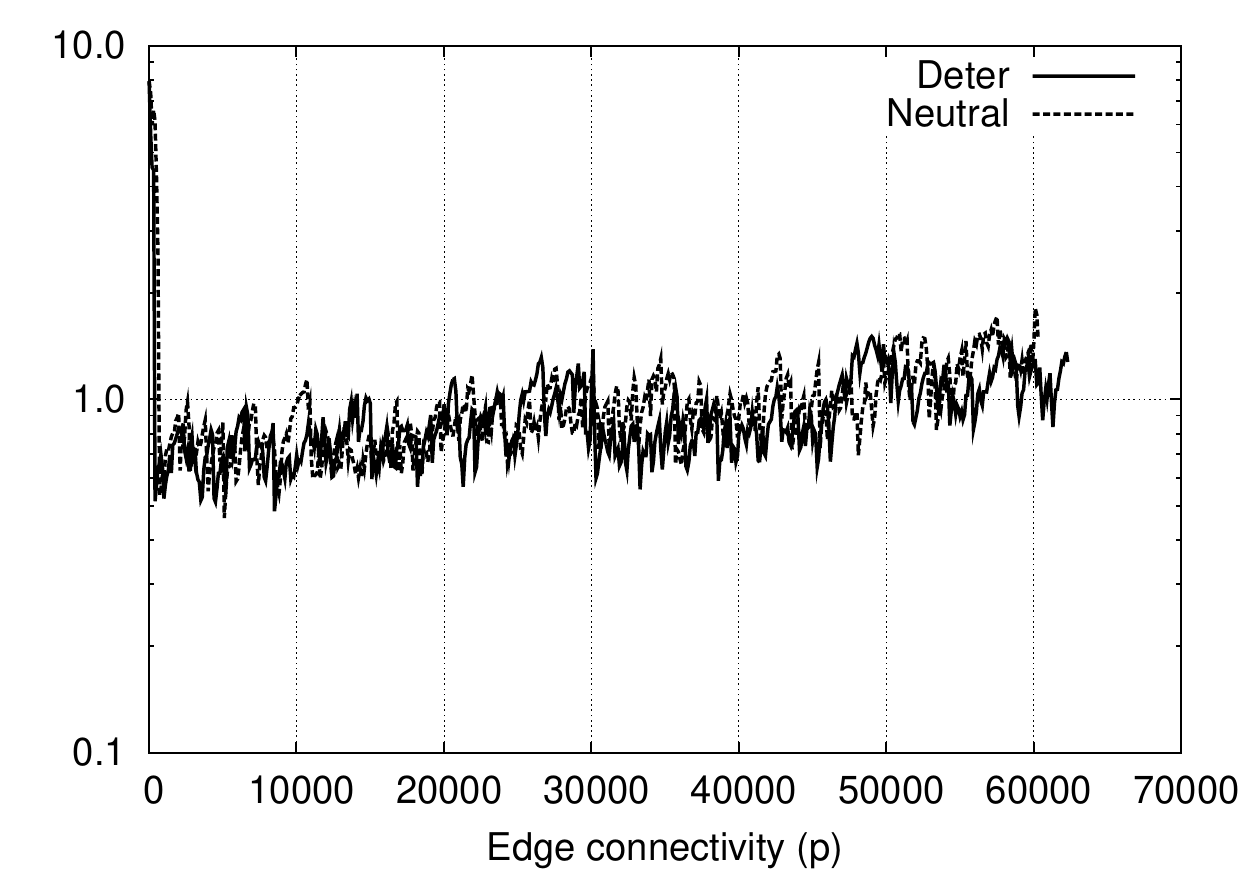}}
\caption{Characteristics of the HSA-EA runs on equi-partite graph with $p=0.010$}
\label{fig:Sub_4}
\end{figure}

As illustrated by Fig.~\ref{fig:Sub_3}.f, all versions of the HSA-EA leaved in average less than $30$ uncolored vertices by the 3-coloring. The bad result by the original HSA-EA coloring the graph with $p=0.0095$ was caused because of the ordering by saturation local search heuristic that got stuck in the local optima. Nevertheless, note that most important measure is $SR$.
\\\\
\textit{The impact of the neutral survivor selection}
\\
In this experiments the impact of the neutral survivor selection on results of the HSA-EA was analyzed. In this context, the HSA-EA with deterministic survivor selection was developed with the following characteristic:
\begin{itemize}
  \item The Equation~\ref{eq:fit1} that prevents the generation of neutral solutions was used instead of the Equation~\ref{eq:fit}.
  \item The deterministic survivor selection was employed instead of the neutral survivor solution. This selection orders the solutions according to the increasing values of the fitness function. In the next generation the first $\mu$ solutions is selected to survive.
\end{itemize}
Before starting with the analysis, we need to prove the existence of neutral solution and to establish they characteristics. Therefore, a typical run of the HSA-EA with neutral survivor selection is compared with the typical run of the HSA-EA with the deterministic survivor selection. As example, the 3-coloring of the equi-partite graph with $p=0.010$ was taken into consideration. This graph is easy to solve by both versions of the HSA-EA. Characteristics of the HSA-EA by solving it are presented in Fig.~\ref{fig:Sub_4}.

In the Fig.~\ref{fig:Sub_4}.a the best and the average number of uncolored nodes that were achieved by the HSA-EA with neutral and the HSA-EA with deterministic survivor selection are presented. The figure shows that the HSA-EA with the neutral survivor selection converge to the optimal solution very fast. To improve the number of uncolored nodes from 140 to 10 only 10,000 solutions to evaluation were needed. After that, the improvement stagnates (slow progress is detected only) until the optimal solution is found. The closer look at the average number of uncolored nodes indicates that this value changed over every generation. Typically, the average fitness value is increased when the new best solution is found because the other solutions in the population try to adapt itself to the best solution. This self-adaptation consists of adjusting the step sizes that from larger starting values becomes smaller and smaller over the generations until the new best solution is found. The exploring of the search space is occurred by this adjusting of the step sizes. Conversely, the average fitness values are changed by the HSA-EA in the situations where the best values are not found as well. The reason for that behavior is the stochastic evaluation function that can evaluate the same permutation of vertices always differently.

More interestingly, the neutral solution occurs when the average fitness values comes near to the best (Fig.~\ref{fig:Sub_4}.b). As illustrated by this figure, the most neutral solutions arise in the later generations when the population becomes matured. In example from Fig.~\ref{fig:Sub_4}.b, the most neutral solutions occurred after 20,000 and 30,000 evaluations of fitness function, where almost $30\%$ of neutral solution occupied the current population.

In contrary, the HSA-EA with deterministic survivor selection starts with the lower number of uncolored vertices (Fig.~\ref{fig:Sub_4}.c) than the HSA-EA with neutral selection. However, the convergence of this algorithm is slower than by its counterpart with the neutral selection. A closer look at the average fitness value uncovers that the average fitness value never come close to the best fitness value. A falling and the rising of the average fitness values are caused by the stochastic evaluation function.

In the Fig.~\ref{fig:Sub_4}.d a diversity of population as produced by the HSA-EA with different survivor selections is presented. The diversity of population is calculated as a standard deviation of the vector consisting of the mean weight values in the population. Both HSA-EA from this figure lose diversity of the initial population (close to value 8.0) very fast. The diversity falls under the value 1.0. Over the generations this diversity is raised until it becomes stable around the value 1.0. Here, the notable differences between curves of both HSA-EA are not observed.

\begin{figure}[H]	
\centering
\subfigure[Heuristic initialization procedure] {\includegraphics[width=6.4cm]{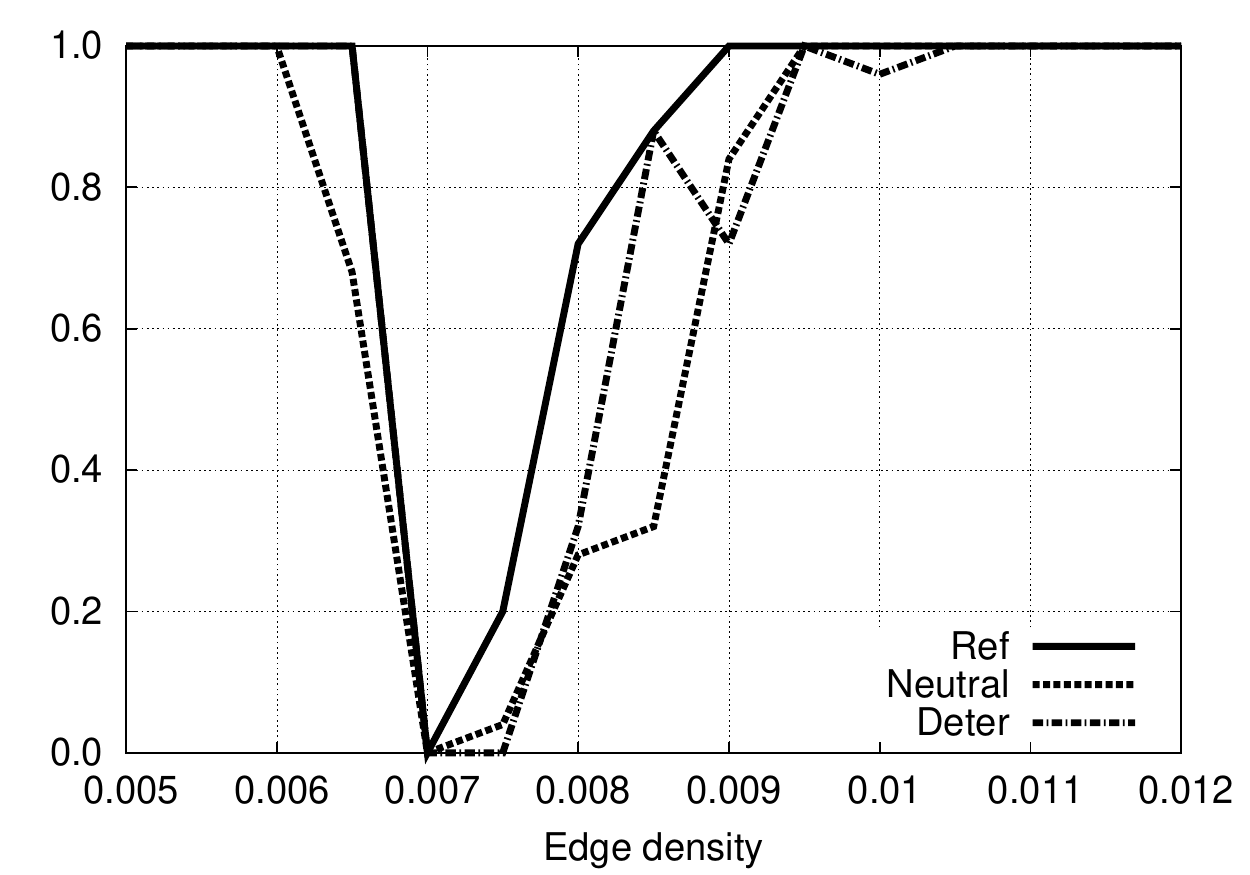}}
\subfigure[Random initialization procedure] {\includegraphics[width=6.4cm]{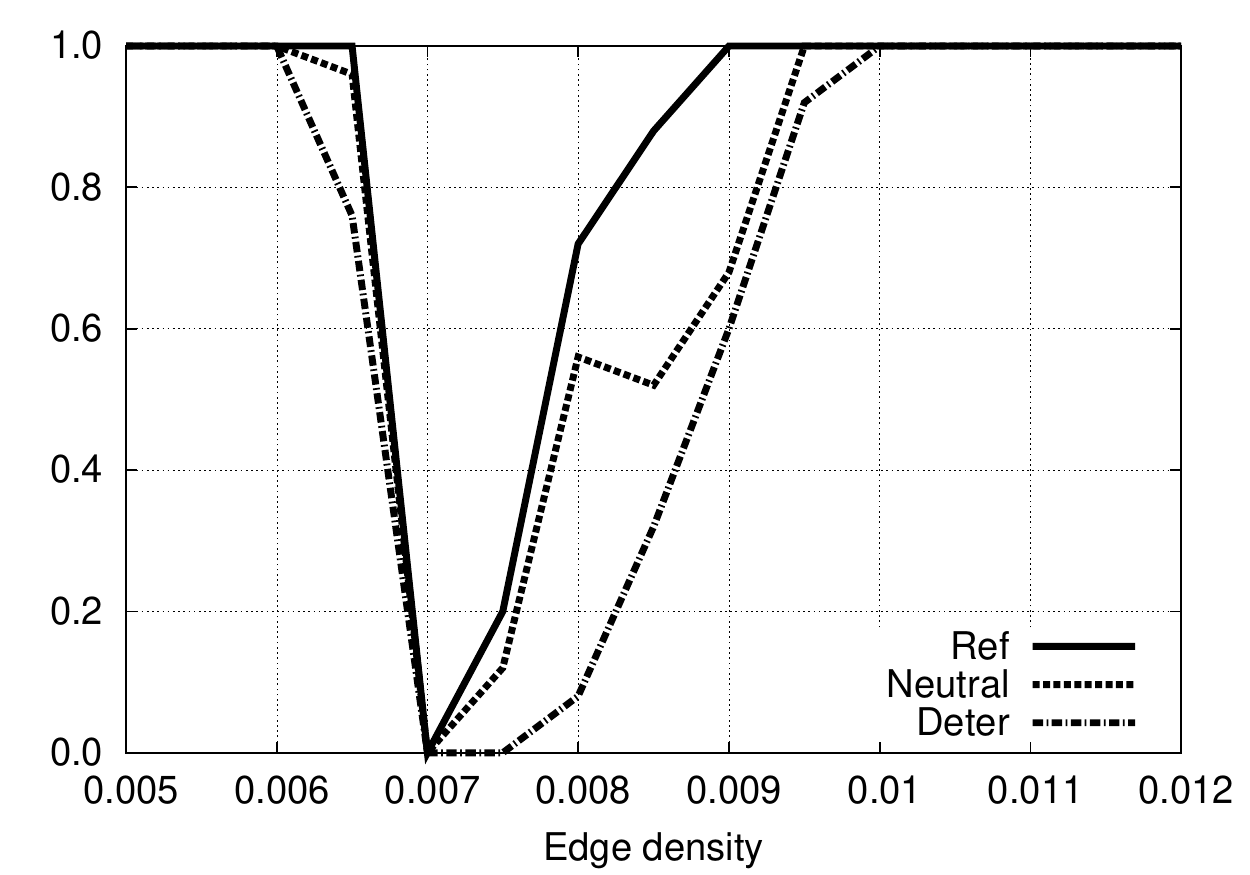}}
\caption{Comparison of the HSA-EA with different survivor selections according to the SR}
\label{fig:Sub_5}
\end{figure}

To determine what impact the neutral survivor selection has on results of the HSA-EA, a comparison between results of the HSA-EA with neutral survivor selection ($Neutral$) and the HSA-EA with deterministic survivor selection ($Deter$) was done. However, both versions of the HSA-EA run without local search heuristics. Results of these  are represented in the Fig.~\ref{fig:Sub_5}. As reference point, the results of the original HSA-EA hybridized with the swap local search heuristic ($Ref$) that obtains the overall best results are added to the figure. The figure is divided in two graphs where the first graph (Fig.~\ref{fig:Sub_5}.a) presents results of the HSA-EA with heuristic initialization procedure and the second graph (Fig.~\ref{fig:Sub_5}.b) results of the HSA-EA with random initialization procedure according to the $SR$.

As shown by the Fig.~\ref{fig:Sub_5}.a the HSA-EA with neutral survivor selection ($Neutral$) exposes better results by the instances near to the threshold ($p \in [0.0075 \ldots 0.008]$) while the HSA-EA with deterministic survivor selection ($Deter$) was slightly better by the instance of graph with $p=0.0085$. Interestingly, while the curve of the former regularly increases the curve of the later is sawing because it raises and falls from the instance to the instance. In contrary, from the Fig.~\ref{fig:Sub_5}.b it can be seen that the HSA-EA with neutral survivor selection outperforms its counterpart with deterministic survivor selection by all instances of random graphs if the random initialization procedure is applied.

In summary, the original HSA-EA with swap local search heuristic used as reference outperforms all observed versions of the HSA-EA.

\subsubsection{Summary}

In this subsection the characteristics of the HSA-EA were studied on the collection of equi-partite graphs, where we focused on the behavior of the algorithm in the vicinity of the threshold. Therefore, an impact of the hybridizing elements, like the initialization procedure, the local search  heuristics, and neutral survivor selection, on results of the HSA-EA are compared. The results of these comparisons in vicinity of the threshold ($p \in [0.0065 \ldots 0.010]$) are presented in Table~\ref{tab:summary}, where these are arranged according to the applied selection (column $Sel.$), the local search heuristics (column $LS$) and initialization procedure (column $Init$). In column $SR$ average results of the corresponding version of the HSA-EA are presented. Additionally, the column $SR_{avg1}$ denotes the averages of the HSA-EA using both kind of initialization procedure. Finally, the column $SR_{avg2}$ represents the average results according to $SR$ that are dependent on the different kind of survivor selection only.

\begin{table}[tbh]
\caption{Average results of various versions of the HSA-EA according to the SR}
\label{tab:summary}
\centering
\begin{tabular}{|l|l|l|c|c|c|}
\hline
Sel. & LS & Init & $SR$ & $SR_{avg1}$ & $SR_{avg2}$ \\
\hline
\multirow{4}{*}{Neut.} & \multirow{2}{*}{No} & $\textnormal{Rand}_{1} $ & 0.52 & \multirow{2}{*}{0.56} & \multirow{4}{*}{0.62} \\
 & & $\textnormal{Heur}_{1} $ & 0.61 & & \\\cline{2-5}
 & \multirow{2}{*}{Yes} & $\textnormal{Rand}_{2} $ & 0.66 & \multirow{2}{*}{0.66} & \\
 & & $\textnormal{Heur}_{2} $ & \textbf{0.67} & & \\\hline
\multirow{4}{*}{Det.} & \multirow{2}{*}{No} & $\textnormal{Rand}_{3} $ & 0.46 & \multirow{2}{*}{0.53} & \multirow{4}{*}{0.57} \\
 & & $\textnormal{Heur}_{3} $ & 0.61 & & \\\cline{2-5}
 & \multirow{2}{*}{Yes} & $\textnormal{Rand}_{4} $ & 0.60 & \multirow{2}{*}{0.60} & \\
 & & $\textnormal{Heur}_{4} $ & 0.61 & & \\\hline
\end{tabular}
\end{table}

As shown by the table~\ref{tab:summary}, results of the HSA-EA with deterministic survivor selection without local search heuristics and without random initialization procedure ($SR=0.46$, denoted as $\textnormal{Rand}_{3}$) were worse than results or its counterpart with neutral survivor selection ($SR=0.52$, denoted as $\textnormal{Rand}_{1}$) in average for more than 10.0\%. Moreover, the local search heuristics improved results of the HSA-EA with neutral survivor selection and random initialization procedure from $SR=0.52$ (denoted as $\textnormal{Rand}_{1}$) to $SR=0.66$ (denoted as $\textnormal{Rand}_{2}$) that amounts to almost 10.0\%. Finally, the heuristic initialization improved results of the HSA-EA with neutral selection and with local search heuristics from $SR=0.66$ (denoted as $\textnormal{Rand}_{2}$) to $SR=0.67$ (denoted as $\textnormal{Heur}_{2}$), i.e. for 1.5\%. Note that the $SR=0.67$ represents the best result that was found during the experimentation.

In summary, the construction heuristics has the most impact on results of the HSA-EA. That is, the basis of the graph 3-coloring represents the self-adaptive evolutionary algorithm with corresponding construction heuristic. However, to improve results of this base algorithm additional hybrid elements were developed. As evident, the local search heuristics improves the base algorithm for $10.0\%$, the neutral survivor selection for another $10.0\%$ and finally the heuristic initialization procedure additionally $1.5\%$.

\section{Conclusion}

Evolutionary algorithms are a good general problem solver but suffer from a lack of domain specific knowledge.
However, the problem specific knowledge can be added to evolutionary algorithms by hybridizing different parts
of evolutionary algorithms. In this chapter, the hybridization of search and selection operators
are discussed. The existing heuristic function that constructs the solution of the problem in a traditional way can be used and
embedded into the evolutionary algorithm that serves as a generator of new solutions.  Moreover, the generation of new solutions can
be improved by local search heuristics, which are problem specific.
To hybridized selection operator a new neutral selection operator has been developed that is capable to deal with neutral solutions, i.e., solutions that
have the different genotype but expose the equal values of objective function. The aim of this operator is to
directs the evolutionary search into new undiscovered regions of the search space,
while on the other hand exploits problem specific knowledge.
To avoid wrong setting of parameters that control the behavior of the evolutionary algorithm, the self-adaptation is used as well.
Such hybrid self-adaptive evolutionary algorithms have been applied to the the graph 3-coloring that is well-known NP-complete problem.
This algorithm was applied to the collection of random graphs, where the phenomenon of a threshold was captured. A threshold determines the instanced of random generated graphs that are hard to color. Extensive experiments shown that this hybridization greatly improves
the results of the evolutionary algorithms. Furthermore, the impact of the particular hybridization is analyzed
in details as well.

In continuation of work the graph $k$-coloring will be investigated. On the other hand, the neutral selection operator needs to be improved with tabu search that will prevent that the reference solution will be selected repeatedly.

\bigskip{\small \smallskip\noindent Updated 1 December 2012.}
\end{document}